\documentclass{SCIS2019}
\usepackage{color}
\usepackage{url}
\usepackage{multirow}
\usepackage{makecell}
\usepackage{array}
\usepackage[table]{xcolor}

\begin{document}
%\oa
%%%%%%%%%%%%%%%%%%%%%%%%%%%%%%%%%%%%%%%%%%%%%%%%%%%%%%%
%%% Authors do not modify the information below
%%% ×÷Õß²»ÐèÒªÐÞ¸Ä´Ë´¦ÐÅÏ¢
\ArticleType{RESEARCH PAPER}
%\SpecialTopic{}
\Year{2019}
\Month{}
\Vol{}
\No{}
\DOI{}
\ArtNo{}
\ReceiveDate{}
\ReviseDate{}
\AcceptDate{}
\OnlineDate{}
%%%%%%%%%%%%%%%%%%%%%%%%%%%%%%%%%%%%%%%%%%%%%%%%%%%%%%%

%%% title: ±êÌâ
%%%   \title{title}{title for citation}
\title{Fewer is More: Efficient Object Detection in Large Aerial Images}{Title for citation}

%%% Corresponding author: Í¨ÐÅ×÷Õß
%%%   \author[number]{Full name}{{email@xxx.com}}
%%% General author: Ò»°ã×÷Õß
%%%   \author[number]{Full name}{}
\author[1]{Xingxing~Xie}{}
\author[1]{Gong~Cheng}{{gcheng@nwpu.edu.cn}}
\author[1]{Qingyang~Li}{}
\author[1]{Shicheng~Miao}{}
\author[2]{Ke~Li}{}
\author[1]{Junwei~Han}{}

%%% Author information for page head. Ò³Ã¼ÖÐµÄ×÷ÕßÐÅÏ¢
\AuthorMark{X. Xie}

%%% Authors for citation. Ê×Ò³ÒýÓÃÖÐµÄ×÷ÕßÐÅÏ¢
\AuthorCitation{X. Xie, G. Cheng, Q. Li, et al}

%%% Authors' contribution. Í¬µÈ¹±Ï×
%\contributions{Authors A and B have the same contribution to this work.}

%%% Address. µØÖ·
%%%   \address[number]{Affiliation, City {\rm Postcode}, Country}
\address[1]{School of Automation, Northwestern Polytechnical University, Xi'an {\rm 710072}, China}
\address[2]{Zhengzhou Institute of Surveying and Mapping, Zhengzhou {\rm 450052}, China}

%%% Abstract. ÕªÒª
\abstract{Current mainstream object detection methods for large aerial images usually divide large images into patches and then exhaustively detect the objects of interest on all patches, no matter whether there exist objects or not. This paradigm, although effective, is inefficient because the detectors have to go through all patches, severely hindering the inference speed. This paper presents an Objectness Activation Network (OAN) to help detectors focus on \textbf{fewer} patches but achieve \textbf{more} efficient inference and \textbf{more} accurate results, enabling a simple and effective solution to object detection in large images. In brief, OAN is a light fully-convolutional network for judging whether each patch contains objects or not, which can be easily integrated into many object detectors and jointly trained with them end-to-end. We extensively evaluate our OAN with five advanced detectors. Using OAN, all five detectors acquire more than 30.0\% speed-up on three large-scale aerial image datasets, meanwhile with consistent accuracy improvements. On extremely large Gaofen-2 images (29200$\times$27620 pixels), our OAN improves the detection speed by 70.5\%. Moreover, we extend our OAN to driving-scene object detection and 4K video object detection, boosting the detection speed by 112.1\% and 75.0\%, respectively, without sacrificing the accuracy. Code is available at \url{https://github.com/Ranchosky/OAN}.}

%%% Keywords. ¹Ø¼ü´Ê

\maketitle

%%%%%%%%%%%%%%%%%%%%%%%%%%%%%%%%%%%%%%%%%%%%%%%%%%%%%%%
%%% The main text. ÕýÎÄ²¿·Ö
%%%%%%%%%%%%%%%%%%%%%%%%%%%%%%%%%%%%%%%%%%%%%%%%%%%%%%%

\begin{figure}[htp]
	\centering
	\begin{minipage}[t]{0.5\textwidth}
		\centering
		\includegraphics[width=0.99\textwidth, height=0.72\textwidth]{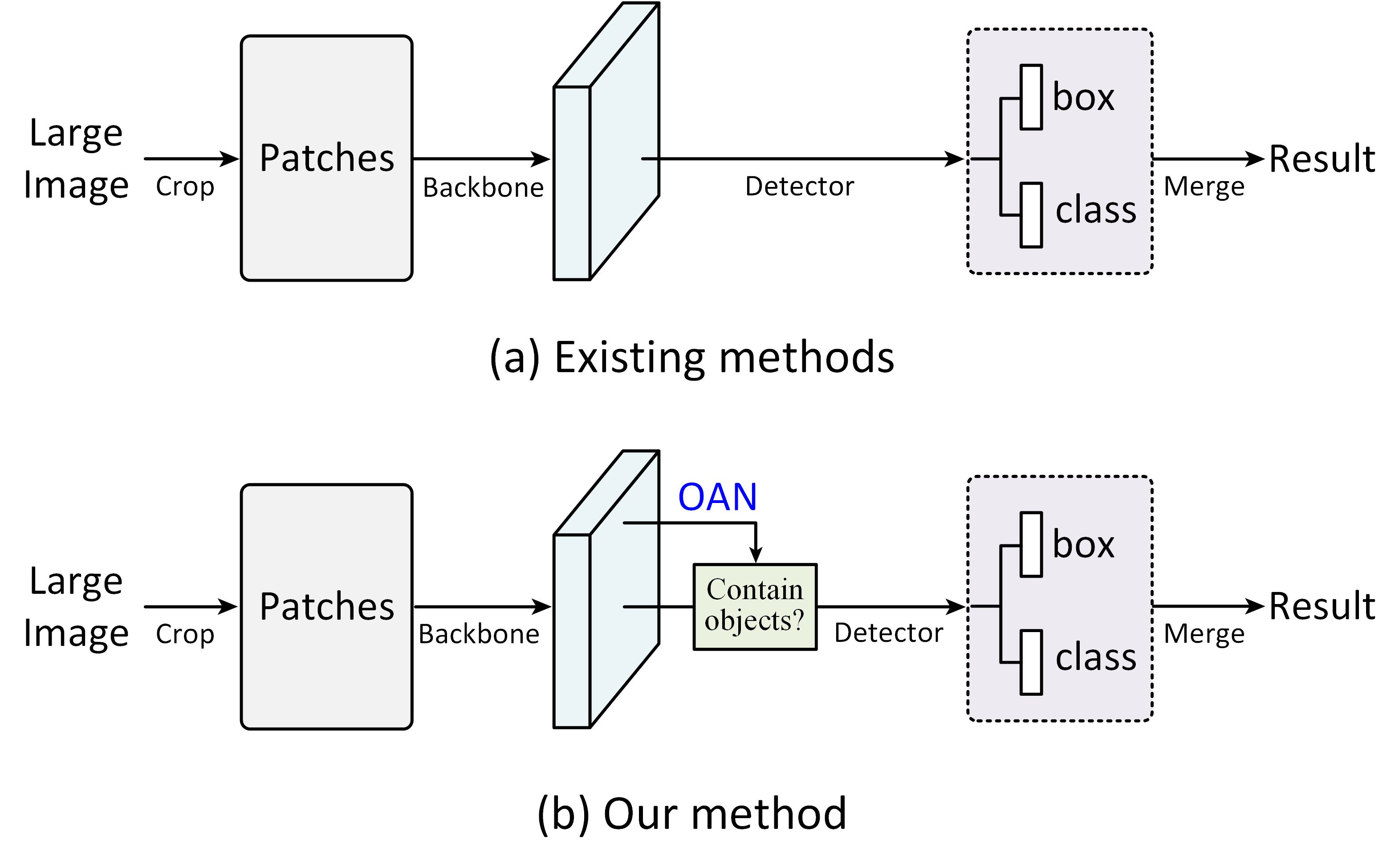}
		\caption{Comparisons of mainstream methods and ours for object detection in large aerial images. (a) Mainstream methods (e.g., RoI Transformer, S2ANet, and Oriented R-CNN) perform detection patch by patch, regardless of whether there are objects or not in each patch. (b) Our method only focuses on the detection of the patches with objects via OAN.}
		%\hspace{5}
		\label{Fig1}
	\end{minipage}
    \hspace{1.7mm}
	\begin{minipage}[t]{0.47\textwidth}
		\centering
		\hspace{-5mm}\includegraphics[width=0.92\textwidth, height=0.80\textwidth]{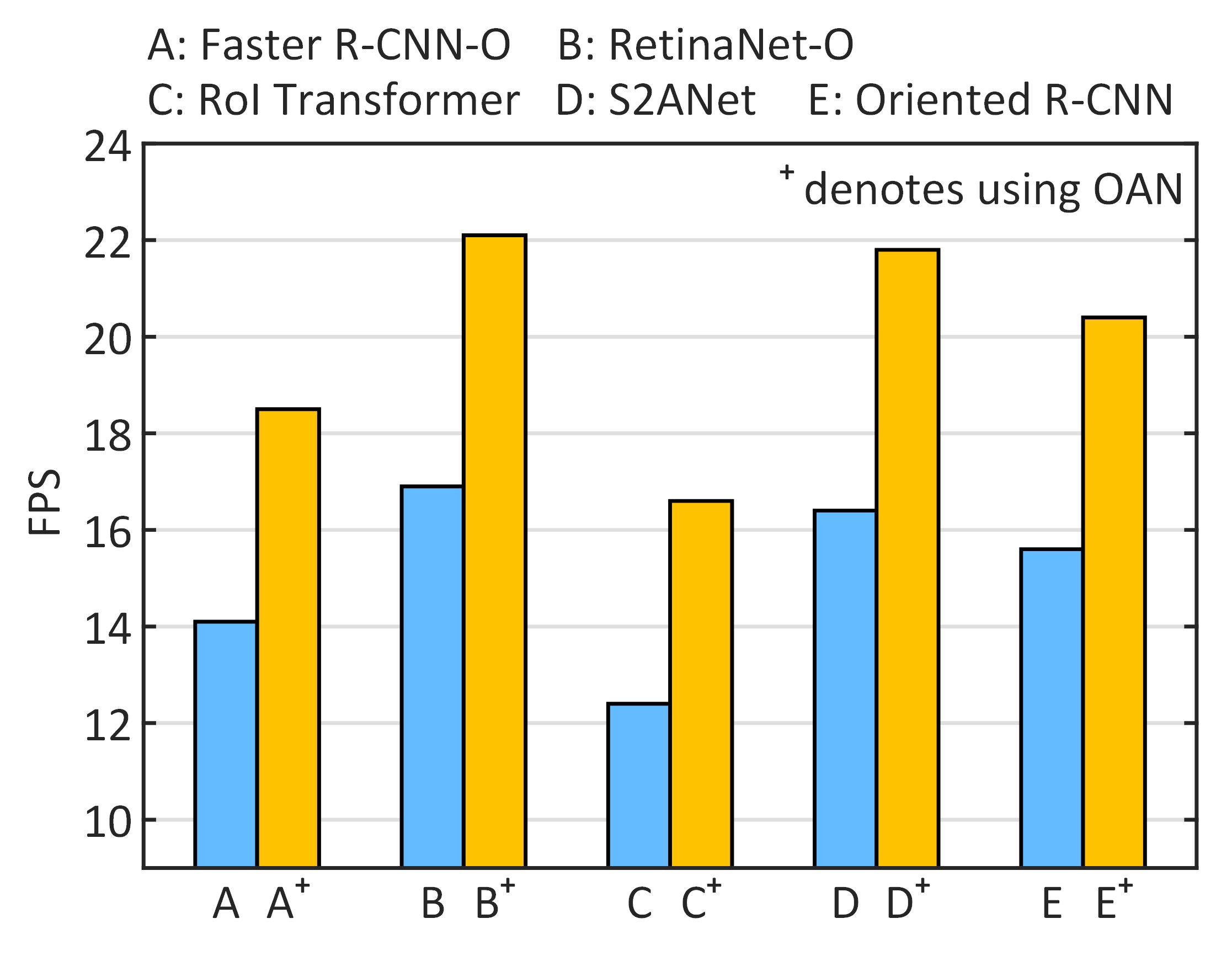}
		\caption{Inference speed comparison between mainstream methods and ours on DOTA-v1.0 test set, using ResNet50 as backbone. Our OAN improves the speed of mainstream methods (e.g., RoI Transformer, S2ANet, and Oriented R-CNN) by more than 30.0\% while with consistent improvements of accuracy (see Table \ref{table1}).}
		\label{Fig2}
	\end{minipage}
\end{figure}

\section{Introduction}
Thanks to the advances in imaging and sensor technologies,  the size of aerial images captured by Unmanned Aerial Vehicles (UAVs) or satellites  has grown significantly (e.g., Gaofen-2 images with 29200$\times$27620 pixels)\cite{ssdrb_tpmi22, dota20_tpami21}. The large aerial images generally contain more information due to the bird's eye imaging and the wider field of view, enabling us to better observe the earth. As one of the key and fundamental tasks of image interpretation, object detection in large aerial images has been an important topic and received extensive studies because of its wide real-world applications.

Driven by the powerful deep learning technologies{\cite{resnet_cvpr16,senet_cvpr18,densnet_cvpr18,hrnet_cvpr19}} and large-scale datasets with annotations{\cite{dota10_cvpr18, dior_ispr20, dota20_tpami21}}, object detection in large aerial images has achieved impressive progress{\cite{roitrans_cvpr19, gv_tpami21, redet_cvpr21, orcnn_iccv21, s2anet_tgrs22, scrdet++_tpami22}} in term of accuracy. However, how to achieve efficient object detection in large aerial images is  notoriously challenging: large aerial images can not be directly fed into the state-of-the-art detectors due to the limitation of Graphics Processing Unit (GPU) memory capacity, and also downsampling large images to the size that detectors can take as input would lose detailed information, especially for small objects. To this end, the existing mainstream object detection methods{\cite{roitrans_cvpr19,s2anet_tgrs22,orcnn_iccv21}} for large aerial images usually work in the following paradigm (see Figure \ref{Fig1}(a)): cropping each large image into massive fixed-size patches (e.g., 1024$\times$1024) by sliding windows with overlapping, then feeding all of them into the detectors, no matter whether there exist objects or not, and finally merging the detection results. 

\begin{figure*}
	\begin{center}
		\includegraphics[width=0.98\linewidth]{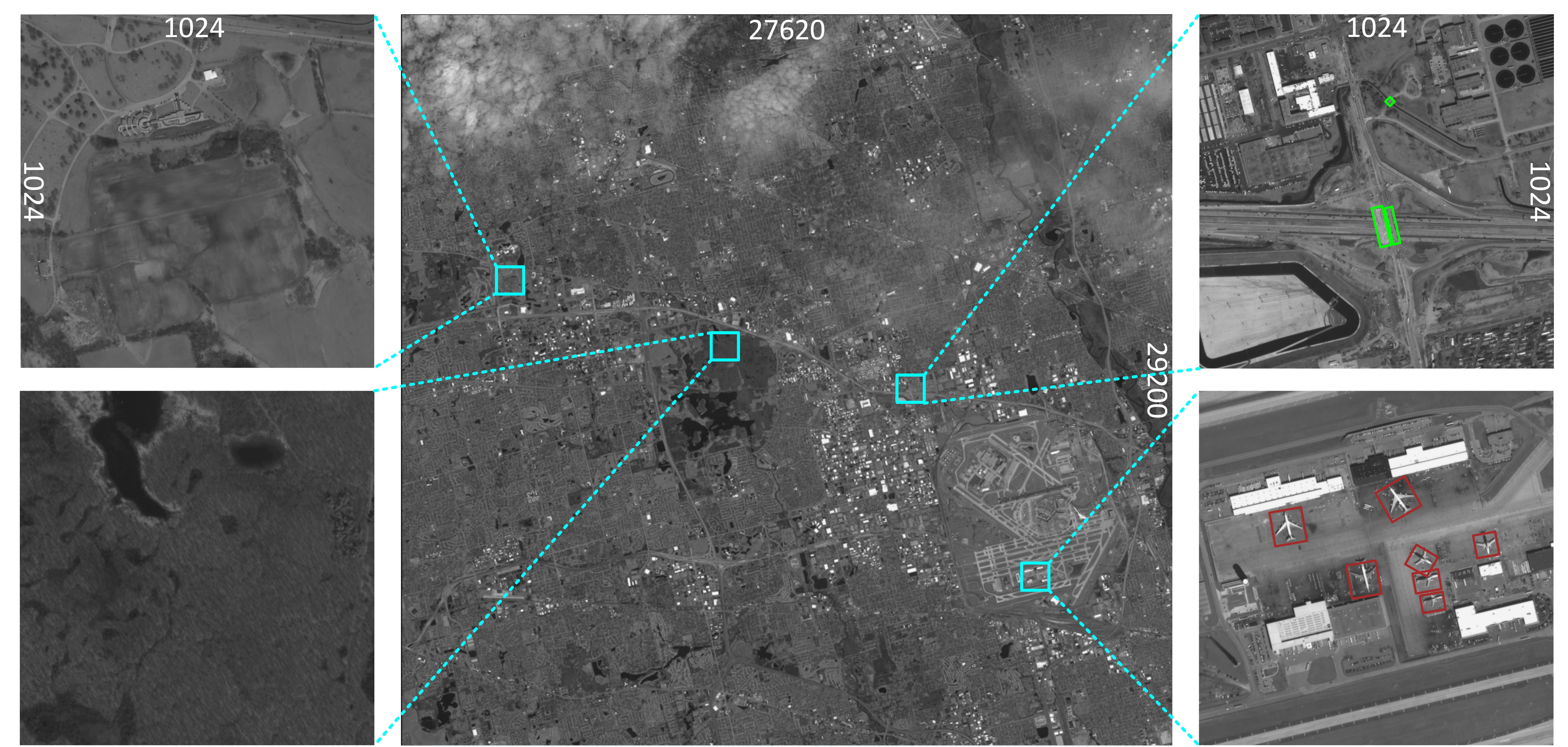}\\
	\end{center}
	\vspace*{-4mm}
	\caption{An example of large aerial image with the size of 29200$\times$27620 pixels (middle). The left and right columns are the enlargements of the corresponding patches (1024$\times$1024) from the large-sized image for better visibility. Red and green boxes are the bounding-box annotations of bridges and airplanes. As seen in the zoomed-in patches, the objects are sparsely distributed and tend to be highly clustered in certain regions, e.g., airplanes generally appear at the airports.}\label{Fig3}
	\vspace{1mm}
\end{figure*}

Nevertheless, the objects in large aerial images are usually sparsely distributed (see Figure \ref{Fig3}) \cite{dota20_tpami21, clusdet_iccv19} and tend to be highly clustered in certain regions, e.g., airplanes generally appear at the airports. As a result, dividing large images into patches will produce numerous patches not containing any objects. Here, for short, we call the patches containing objects as \emph{valid patches}, and the ones without objects as \emph{invalid patches}. For the current mainstream object detection methods, all the \emph{invalid patches} will pass through the detectors, thus consuming a lot of computation resources and severely restricting the efficiency of detection. On the other hand, numerous \emph{invalid patches} could increase the risk of false alarms and degrade the detection accuracy. Under this circumstance, a natural question to ask is: can we let detectors only focus on \textbf{fewer} \emph{valid patches} to achieve \textbf{more} efficient inference and \textbf{more} accurate detection in an elegant way? In fact, the key to the question lies in how to confidently judge whether each patch contains objects or not, i.e., the objectness prediction for each patch.

In this work, we propose an Objectness Activation Network, termed OAN, which is conceptually simple yet highly effective for measuring whether each patch includes the objects of interest. In brief, OAN is built on the convolutional feature maps from the last stage of the backbone that can be any kind of commonly-used deep  feature extraction networks. It adopts grid-wise strategy to divide each image patch into a number of grids (e.g., 16$\times$16), and predicts the objectness score at each grid by adding a few convolutional layers. Thanks to the surprisingly simple design, OAN can be easily applied to many detectors for boosting detection efficiency and jointly trained end-to-end (see Figure \ref{Fig1}(b)). 

We extensively evaluate our OAN with five advanced detectors. With the help of OAN, all five detectors acquire more than 30.0\% speed-up (see Figure \ref{Fig2}) on three large-scale aerial image datasets, including DOTA-v1.0, DOTA-v1.5, and DOTA-v2.0. Surprisingly, our OAN also boosts the detection accuracy due to the elimination of false alarms from \emph{invalid patches}. Typically, on DOTA-v1.0, with the patch size of 1024$\times$1024, (i) using Oriented R-CNN, our OAN achieves the highest accuracy of 76.73\% mAP running at 17.5 FPS (30.5\% speed gain); (ii) employing S2ANet, our OAN achieves 74.37\% mAP running at 21.8 FPS (32.9\% speed gain). Both the accuracy and speed of our OAN outperform all existing detectors. In addition, on extremely large Gaofen-2 images (29200$\times$27620 pixels), combining Oriented R-CNN with OAN significantly improves the inference speed by 70.5\% but without sacrificing the detection accuracy. 

Moreover, we showcase the generality of our OAN via the task of driving-scene object detection and 4K video object detection. Specifically, on the large-scale driving-scene dataset, integrating OAN into two popular baseline detectors, namely the Faster R-CNN~\cite{faster_tpami17} and RetinaNet~\cite{retinanet_iccv17}, could obtain 112.1\% and 98.8\% speed-up, respectively. For the task of object detection in 4K videos, our OAN still achieves 75.0\% and 70.4\% speed gains compared with the baseline detectors. Besides acquiring considerable speed-up, our OAN also improves the detection accuracy of both tasks under stricter metrics.   

In summary, OAN is quite simple yet extremely effective. It makes the detectors focus on \textbf{fewer} patches to achieve \textbf{more} efficient inference speed and \textbf{more} accurate results. We hope our simple yet effective method will serve as a general baseline while facilitating future research for real-world applications.

\section{Related work}
\subsection{Object Detection in Aerial Images}
Driven by convolutional neural networks (CNNs) and advanced detection frameworks {\cite{faster_tpami17, retinanet_iccv17, cascadercnn_cvpr18, cornernet_eccv18, fcos_iccv19, atss_cvpr20, zhao2020gtnet, tood_iccv21, aod, qbox, wang2022gatector, canet,od_survey}}, object detection in aerial images~\cite{ricnn_tgrs16,aol_tgrs17,rifd_tip19, roitrans_cvpr19, orcnn_iccv21,wang2021multiple,fen,DODet} has achieved promising results, especially in the directions of rotation-invariant object detection {\cite{ricnn_tgrs16, aol_tgrs17, rifd_tip19}} and oriented object detection{\cite{roitrans_cvpr19, csl_eccv20, gv_tpami21,redet_cvpr21, dcl_cvpr21, aopg_tgrs22, scrdet++_tpami22}}. 

To address the challenge that the objects in aerial images usually have big variations in orientation, many works{\cite{ricnn_tgrs16, aol_tgrs17, rifd_tip19}} attempted to design methods suitable for aerial images based on general detectors{\cite{faster_tpami17, retinanet_iccv17, fcos_iccv19, qbox, ji2019small}}. Representatively, RICNN{\cite{ricnn_tgrs16}} explores the extraction of rotation-invariant features, which adds a fully-connected layer for learning rotation-invariant features on R-CNN framework.  

Since the release of large-scale aerial image dataset DOTA{\cite{dota10_cvpr18}}, oriented object detection in aerial images has attracted particular attention and many notable approaches have been proposed {\cite{roitrans_cvpr19,s2anet_tgrs22, redet_cvpr21, orcnn_iccv21, kld_nips21}}. For instance, Ding et al.{\cite{roitrans_cvpr19}} proposed RoI transformer for generating high-quality oriented proposals, which greatly improves the detection accuracy of oriented objects. The work of{\cite{s2anet_tgrs22}} designs a one-stage oriented object detection network, termed S2ANet, by introducing feature alignment operation and regression refinement on RetinaNet{\cite{retinanet_iccv17}}. Oriented R-CNN{\cite{orcnn_iccv21}} presents an effective oriented proposal generation network, which makes two-stage detector achieve faster detection speed and higher accuracy.

As we know, the size of aerial images is usually very large. Due to the GPU memory capacity, it is incapable to directly feed the whole images into detectors. To deal with this issue, most of the current mainstream detection methods often partition the original large images into uniform patches, and then exhaustively detect the objects of interest on all of them, no matter whether there exist objects or not. Obviously, this exhaustive paradigm is incompetent and it is absolutely imperative to explore efficient object detection solution to large aerial images, which is also the motivation of this work.

\subsection{ Speeding Up Detection in Large Images}
With the development of imaging techniques, we are facing more and more images with large sizes, such as satellite images, UAV images, driving-scene images, 4K/8K videos, to name a few. 
How to effectively detect objects in large images has gained an increasing amount of attention. 

The straightforward approaches to speeding up detection are to design efficient network architectures, such as YOLO~\cite{yolo_cvpr16}, RetinaNet~\cite{retinanet_iccv17}, RefineDet~\cite{refinedet}, HSD~\cite{hsd}, and FCOS~\cite{fcos_iccv19}. These high-efficiency detectors do accelerate detection to some extent, whereas the computation cost increases dramatically when directly applying them to large images.

More recently, some exploratory works have been proposed \cite{ass_cvpr15, clusternet_cvpr18, dzn_cvpr18, clusdet_iccv19, r2cnn_tgrs19, dmnet_cvprw20,eod_wacv20} to improve the detection speed in large images by the coarse-to-fine cascade structure. 
For example, $R^2$-CNN{\cite{r2cnn_tgrs19}} uses a naive binary classification branch to predict the existence of objects in each patch, thus avoiding the detection of numerous patches not containing any targets. ClusDet {\cite{clusdet_iccv19}} designs a cluster network to generate the candidate object regions in large images and then performs detection selectively. DMNet{\cite{dmnet_cvprw20}} generates the object proposal regions via density map generation network and then conducts detection in these regions. AutoFocus~\cite{autof} adds an extra branch to select the regions that may contain objects, and then crops them from the input image for further detection. CornerNet-Lite~\cite{clite} predicts a set of possible object locations with the attention maps and then performs detection on each region. Moreover, some approaches\cite{dzn_cvpr18,eod_wacv20} introduced reinforcement learning to sequentially select regions for detection.

However, the aforementioned approaches are far from satisfactory in terms of speed-up, training simplicity, and applicability. Taking $R^2$-CNN as an example, which is also the most related work to our OAN, it directly judges if each patch includes objects by a binary classifier, but it has a defect. That is, it often loses numerous patches containing small-sized objects, thus significantly degrading the detection accuracy. Meanwhile, $R^2$-CNN also needs to use \emph{invalid patches} as training samples, which makes the training inconsistent with the detectors as well as expanding the training time. ClusDet and DMNet adopt complex networks (i.e., proposal clustering and density map estimation) to generate regions with objects. They are difficult to be jointly optimized, and their performances are much sensitive to the proposal clustering and density map estimation. CornerNet-Lite downsamples the input image during the region generation, which may lead to the discarding of some regions with small-sized objects. AutoFocus needs to go through multiple merge operations to produce regions during the inference stage, which introduces multiple super-parameters and hinders its speed gain. The reinforcement learning-based approaches depend on intricate optimization strategies to obtain potential regions. They need to design reliable cost functions and undergo time-consuming training.

Different from $R^2$-CNN,  we introduce a simple grid-wise prediction manner and adopt the maximum value of the scores of all grids as the objectness metric for each patch, rather than based on a naive binary classifier, thus effectively retaining as many objects as possible. Actually, $R^2$-CNN is a special case of OAN when we regard each patch (1024$\times$1024) as a grid. In contrast to ClusDet, DMNet, AutoFocus, CornerNet-Lite and reinforcement learning-based approaches, our OAN is a light-weight fully-convolutional network jointly trained with the detectors end-to-end, being very simple yet surprisingly effective. The detailed comparison with those related works can be found in the ablation studies.

\section{Objectness Activation Network}
\begin{figure}[]
	\centering
	\begin{minipage}[]{0.50\textwidth}
		\centering
		\vspace{-21mm}
		\hspace{-1.5mm}\includegraphics[width=0.98\textwidth,height=0.45\textwidth]{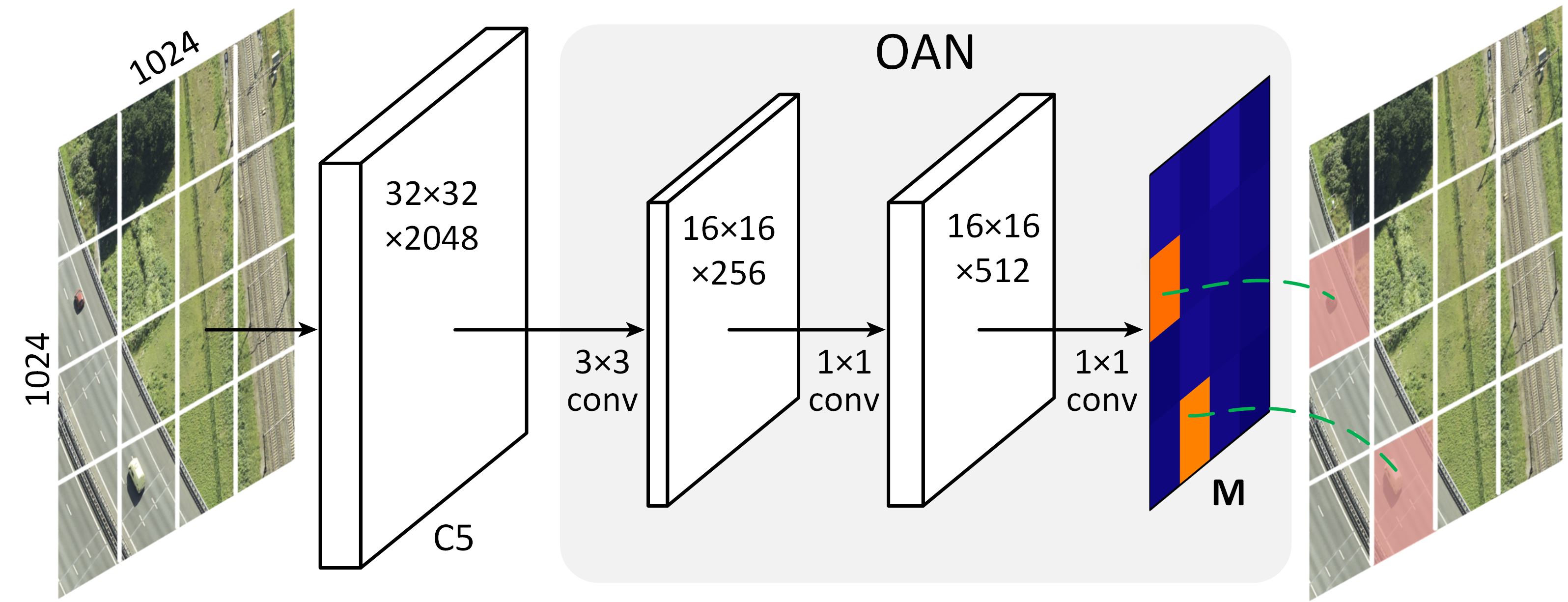} 
		\caption{OAN architecture. It is a fully-convolutional network built on the backbone. Here, C5 denotes the feature maps of the last stage of ResNet50. All the numbers are computed with a 1024$\times$1024 input patch.}
		\label{Fig4}
	\end{minipage}
    \hspace{3mm}
	\begin{minipage}[t]{0.46\textwidth}
		\centering
		\hspace{1mm}\includegraphics[width=0.96\textwidth, height=0.51\textwidth]{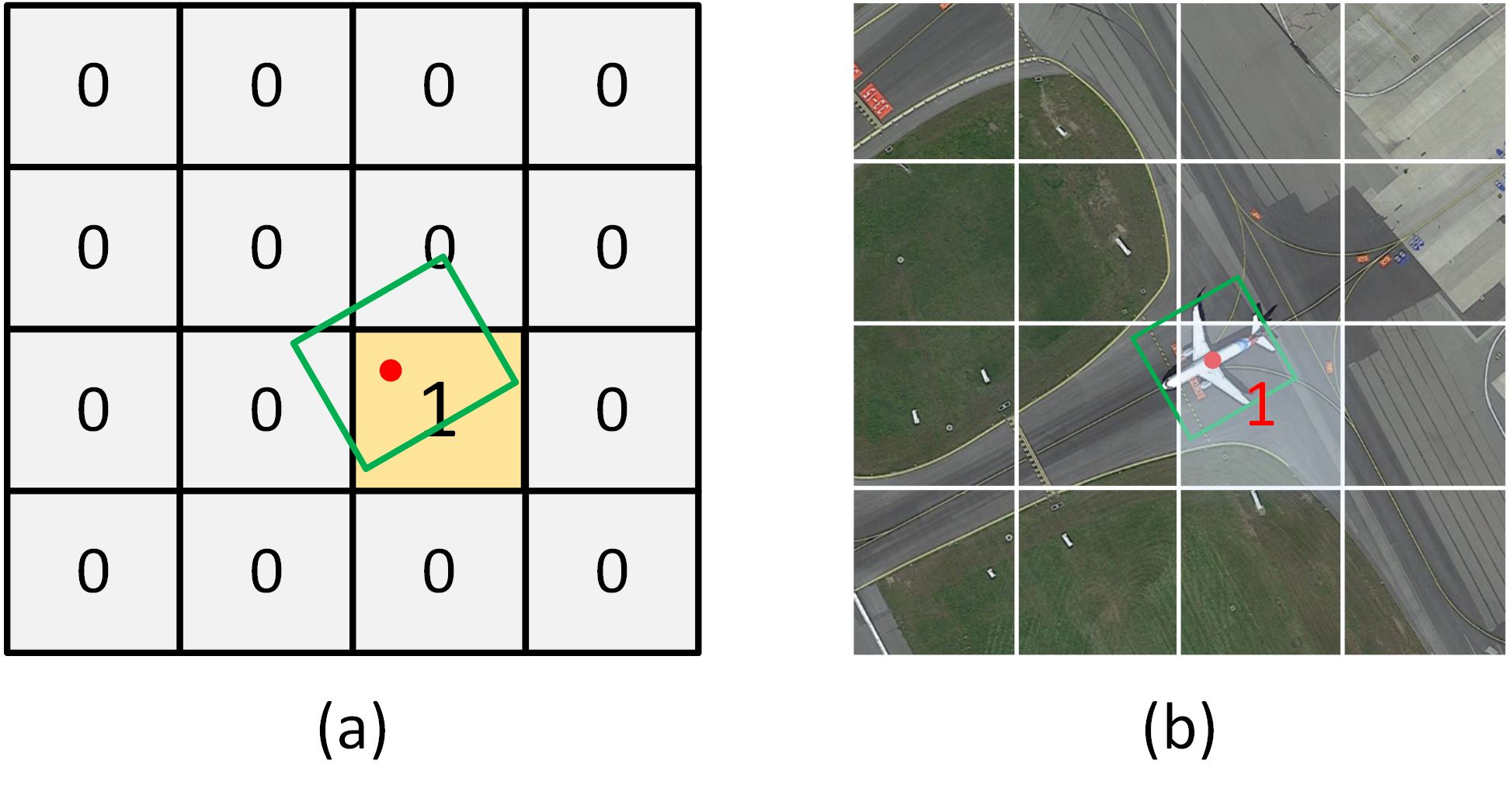}
		\vspace{-2.2mm}
		\caption{Illustration of sample assignment. The green box is the ground-truth bounding box. (a) The definitions of positive and negative samples (1 for positive and 0 for negatives). (b) An example of sample assignment.}
		\label{Fig5}
	\end{minipage}
\end{figure}

As the name says, the Objectness Activation Network (OAN) is a fully-convolutional network (see Figure \ref{Fig4}) for measuring the objectness confidence of each patch, in order to let the detectors focus on the areas where to detect the objects of interest. It takes the patches cropped from large aerial images as input and outputs their corresponding objectness activation maps. Next, we introduce OAN in detail.

\subsection{Network Architecture}

We build OAN on general backbones, such as ResNet{\cite{resnet_cvpr16}}, ResNeXt{\cite{resnxt_cvpr17}}, ResNeSt{\cite{resnest_arxiv2020}}, and Swin Transformer{\cite{swin_iccv21}}, by adding a few convolutional layers to output the objectness scores of each patch. The architecture of OAN is illustrated in Figure \ref{Fig4} and we only instantiate OAN with ResNet50 for clarity. 

Specifically, given an image patch with the size of 1024$\times$1024, we divide it into \emph{S}$\times$\emph{S} grids (\emph{S} is 16 for 1024$\times$1024 input patch and the selection of $S$ will be discussed in the ablation studies). We choose the feature maps extracted from the last stage of ResNet50, C5 for short, as the input of OAN. C5 has a stride of 32 pixels \emph{w.r.t} the input patch. On C5, we first apply a 3$\times$3 stride-2 convolutional layer with 256 filters to obtain a lower-dimensional feature map with the size of 16$\times$16$\times$256. Then we feed it to a 1$\times$1 convolutional layer with 512 filters, followed by another 1$\times$1 convolutional layer with 1 filter to generate the objectness activation map $\mathbf{M}$, with the size of 16$\times$16$\times$1. The value of $\mathbf{M}$ at each position denotes the objectness score of its corresponding grid.  

As stated above, our OAN has a straightforward structure, which facilitates the integration and practical usage. Note that more complex designs (e.g., multi-scale feature fusion) have potential to further improve the performance but are not the primary goal of this work.

\subsection{Loss Function}\label{section3-1}
Let \{$GT$\} denote the ground-truth boxes of the objects for an image patch $\mathbf{I}$, and $\mathbf{M}$($i$,$j$) be the objectness score of the grid $\mathbf{I}$($i$,$j$), where  $0 \leq i, j \leq S-1$ are the grid indexes.

With these definitions, we design the rule of sample assignment (see Figure \ref{Fig5}). For each grid  $\mathbf{I}$($i$,$j$), we assign a binary label  $\hat{\mathbf{P}}(i, j)$. To be specific, if the center of the \textit{l}-th ground-truth box $GT_{l}$ falls into a grid, the grid is defined as positive, otherwise, the grid is negative. The process of sample assignment only relies on the object position and so is class-independent, which does not involve the competition among classes and reduces the difficulty of optimization for OAN. Meanwhile, unlike works \cite{clusdet_iccv19, dmnet_cvprw20}, the process does not need any additional ground-truth information except the ground-truth boxes. Thus, we formulate the
process of sample assignment as follows:
\begin{equation}
	\hat{\mathbf{P}}(i, j)= \begin{cases}1 & \text { if } GT_{l}(x, y) \in \mathbf{I}(i, j) \\ 0 & \text { otherwise}\end{cases}
	,
\end{equation}
where $GT_{l}(x, y)$ denotes the center of the \textit{l}-th ground-truth box $GT_{l}$. And the loss function of OAN is defined as follows:
\begin{equation}
	\label{equ2}
	L_{\text {OAN }}=\frac{1}{S^{2}} \sum_{i=0}^{S-1} \sum_{j=0}^{S-1} F L(\hat{\mathbf{P}}(i, j), \mathbf{M}(i, j)).
\end{equation}
Here, $FL(\cdot)$ is the conventional focal loss{\cite{retinanet_iccv17}} to balance negative and positive samples. Obviously, the sample assignment and the loss function of OAN are simple, thus well facilitating fast training.

\begin{figure*}
	\begin{center}
		\includegraphics[width=0.98\linewidth]{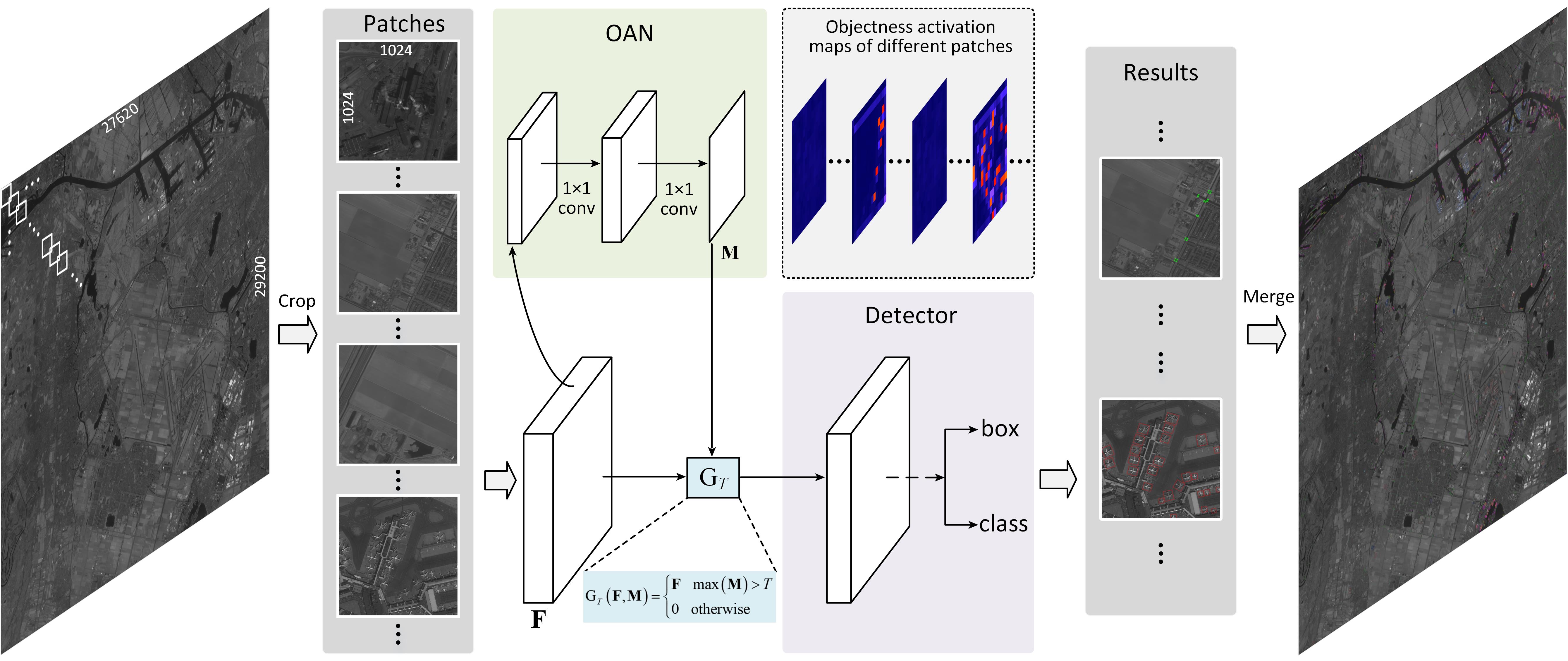}\\
	\end{center}
	\vspace*{-3mm}
	\caption{The unified detection framework for large aerial images. $\mathbf{F}$ stands for the feature maps output by the backbone. OAN serves as the role of generating objectness activation maps of input patches. The gate  $\mathrm{G}_{T}$ is used to decide whether to perform detection for the current input patch. As for the detector, it can be one-stage or two-stage (e.g., RoI Transformer, S2ANet, Oriented R-CNN, to name a few).}\label{Fig6}
	\vspace{1mm}
\end{figure*}
\section{OAN for Object Detection}
For OAN, our ultimate goal is to jointly optimize it with the detectors. Thus far, we have described the its architecture and loss function without considering how to utilize it for object detection in large images. Next, we will introduce in detail how to integrate our OAN with the detectors with shared backbone and optimize them end-to-end by multi-task loss, as shown in Figure \ref{Fig6}.

\subsection{Training}

The overall loss $L$ is the sum of OAN loss, classification loss, and bounding-box loss. Specifically, the training losses of two-stage detectors with OAN and one-stage ones with OAN are defined as 

\begin{equation}
	\label{equ3}
	L=L_{\mathrm{RPN}}+L_{\mathrm{box}}+L_{\mathrm{class}}+\lambda L_{\mathrm{OAN}}, 
\end{equation}
\begin{equation}
	\label{equ4}
	L=L_{\text {box }}+L_{\text {class }}+\lambda L_{\text {OAN }}. 
\end{equation}
\noindent Here, $L_{\mathrm{RPN}}$ is the proposal generation loss, which is specific to proposal-driven detection networks. For more details of  $L_{\mathrm{RPN}}$, we refer readers to {\cite{roitrans_cvpr19,orcnn_iccv21,faster_tpami17}}. $L_{\text {class}}$ and $L_{\text {box}}$ denote the classification loss and regression loss, respectively. Their definitions are identical to {\cite{s2anet_tgrs22,retinanet_iccv17}}. $L_{\text {OAN}}$ is the OAN loss defined in Eq. (\ref{equ2}). $\lambda$  is a trade-off parameter for balancing the losses. With the overall loss $L$, OAN and the detectors can be optimized simultaneously to generate objectness activation maps and then obtain the detection results with OAN-guided detectors.

\subsection{Inference} 
Given an original large image, we first partition it into the patches with the same size of 1024$\times$1024 and any adjacent patches have 200-pixel overlap. Then we feed them into the backbone to extract feature maps $\mathbf{F}$. Based on $\mathbf{F}$, OAN produces the objectness activation map  $\mathbf{M}$ of the corresponding input patch.  It is worth noting that, for the detectors using  Feature Pyramid Network (FPN)~\cite{fpn_cvpr17}, the input of OAN is still the feature maps from the last stage of backbone rather than the FPN's feature maps. We choose the maximum value of $\mathbf{M}$ as the confidence to indicate whether there are objects in each patch or not. If the confidence is higher than the activation threshold $T$, the detector conducts detection for the input patch, otherwise not. We realize this by a gate $\mathrm{G}_{T}$, which is defined as
\begin{equation}
	\mathrm{G}_{T}(\mathbf{F}, \mathbf{M})= \begin{cases}\mathbf{F} & \max (\mathbf{M})>T \\ 0 & \text { otherwise}\end{cases}
	.
\end{equation}

The outputs $\mathbf{O}=\{ class, box \}$ of the detector for each patch is calculated by $f_{\Theta}\left(\mathrm{G}_{T}(\mathbf{F}, \mathbf{M})\right)$, where $f_{\Theta}$ denotes the detection network with parameter $\Theta$. Note that $f_{\Theta}$ can be any detectors, such as RoI Transformer, S2ANet,  Oriented R-CNN, to name a few. Finally, we merge the outputs of the patches and obtain the results on the large images.

\subsection{Implementation Details}
When merging OAN and the detectors into a unified framework, we keep the architecture and the hyper-parameters of the detectors consistent with their source codes. Besides the hyper-parameters of the detectors, there exists an important hyper-parameter, i.e., the activation threshold $T$. It determines the number of patches passing to the detectors, thus affecting the final detection speed and accuracy. Therefore, how to set a reasonable activation threshold is extremely critical for achieving more efficient and more accurate detection. In this paper, we propose a statistic-based method to adaptively obtain the activation threshold, which can be formulated as
\begin{equation}
	\label{equ6}
	\left\{\begin{array}{l}
		T=\frac{1}{k}(m+v)^{2} \\
		v=\operatorname{mean}\left(\operatorname{std} \left(\mathbf{M}_{1}\right), \cdots, \operatorname{std} \left(\mathbf{M}_{2000}\right)\right) \\
		m=\operatorname{mean}\left(\max \left(\mathbf{M}_{1}\right), \cdots, \max \left(\mathbf{M}_{2000}\right)\right)
	\end{array}\right.
	,
\end{equation}
where $k$ is a scaling factor ($k$=4 in this work), $\mathbf{M}_{1}, \mathbf{M}_{2}, \ldots, \mathbf{M}_{2000}$  denote the objectness activation maps of the last 2000 iterations during training, $v$ is the mean of the standard deviation of 2000 objectness activation maps, and $m$ stands for the mean of 2000 objectness activation maps. Here, we choose the last 2000 iterations because the model tends to be stable during this time and the objectness activation maps are more representative. One can also adjust the activation threshold freely based on practical scenarios.

As shown in Eq. (\ref{equ3}) and Eq. (\ref{equ4}), $\lambda$ is another crucial parameter. Since the current research mainly focuses on oriented object detection, we verify our OAN with five popular oriented detectors, namely, RetinaNet-O{\cite{retinanet_iccv17}}, Faster R-CNN-O{\cite{faster_tpami17}}, RoI Transformer{\cite{roitrans_cvpr19}}, S2ANet{\cite{s2anet_tgrs22}} and Oriented R-CNN{\cite{orcnn_iccv21}}. We set the trade-off parameter $\lambda$ for those detectors to 3, 5, 6, 4, 8, respectively, to balance the losses of different terms. Please refer to our source code for more details. Here, RetinaNet-O{\cite{retinanet_iccv17}} and Faster R-CNN-O{\cite{faster_tpami17}} are realized by adding an additional angle prediction branch. 

\section{Experiments}
%We use three large-scale aerial image benchmarks, including DOTA-v1.0, DOTA-v1.5, and DOTA-v2.0, to comprehensively evaluate our method.

\subsection{Datasets and Parameter Settings}\label{section5-1}
\textbf{Datasets: }DOTA-v1.0{\cite{dota10_cvpr18}} is a large-scale dataset for object detection in aerial images. It consists of 2806 images and 188282 instances, covered by 15 object classes: Plane (PL), Helicopter (HC), Ship (SH), Bridge (BR), Harbor (HA), Large vehicle (LV), Small vehicle (SV), Baseball diamond (BD), Ground track field (GTF), Tennis court (TC), Basketball court (BC), Soccer-ball field (SBF), Roundabout (RA), Swimming pool(SP), and Storage tank (ST). The image size ranges from 800$\times$800 to 4000$\times$4000. DOTA-v1.5 uses the same images as DOTA-v1.0, but it adds the annotations of small-sized instances and the Container Crane (CC) class.

DOTA-v2.0{\cite{dota20_tpami21}} is a more large-scale dataset. It has 11268 images and 1793658 instances. Compared with DOTA-v1.5, DOTA-v2.0 adds two new categories: Airport (Air) and Helipad (Heli). To approach the object distribution in real-world applications, DOTA-v2.0 collects more large images (e.g., 29200$\times$27620 Gaofen-2 satellite images and 7360$\times$4912 CycloMedia airborne images). 

As same as previous works{\cite{roitrans_cvpr19,s2anet_tgrs22,orcnn_iccv21,redet_cvpr21}}, we use the training and validation sets for training, and report the mean average precision (mAP) on testing set (no public labels) after submitting the results to evaluation server.

\textbf{Parameter settings: }Following common practice, we divide all original images into 1024$\times$1024 patches with the stride of 824. For testing, we first map the results of all \emph{valid patches} to original large images, and then apply Non-Maximum Suppression (NMS) to these results. The threshold of NMS is set to 0.1. We implement our method based on the OBBDetection\footnote{\url{https://github.com/jbwang1997/OBBDetection}} , which is an open-source Toolkit for oriented object detection. All models are trained and tested with a single Tesla V100 GPU. The batch sizes of training and testing are set to 4 and 1, respectively. We set the initial learning rate to 0.01 and then decrease it by a factor of 10 after 8 epoch and 11 epoch. Consistent with existing work, we use SGD as the optimizer for model training. Unless explicitly specified, we apply  1$\times$ learning schedule (12 epochs) during training by following the setting of \cite{mmdetection_arxiv19}. Moreover, for convenience and fair comparison, we adopt the same metric  of inference speed (FPS) as {\cite{roitrans_cvpr19,s2anet_tgrs22, orcnn_iccv21, orrep_cvpr22}}, that is, processing all 1024$\times$1024 patches cropped from original big images and then calculating the average inference speed.

\subsection{Main Results}
We evaluate our OAN using five mainstream detectors on three large-scale aerial image datasets, namely, DOTA-v1.0, DOTA-v1.5, and DOTA-v2.0 datasets. The source codes of these detectors are publicly available. In the following Tables, H-104, DLA-34, Re50, and RXt50 denote the 104-layer hourglass network{\cite{hourglass_cvpr17}}, 34-layer deep layer aggregation network{\cite{dal_arxiv2019}}, rotation-equivariant ResNet50{\cite{redet_cvpr21}}, and ResNeXt50{\cite{resnxt_cvpr17}}, respectively. R50/101/152 stands for ResNet50/101/152. Following common practices, all detectors are based on FPN {\cite{fpn_cvpr17}} if not particularly indicated.

\begin{table*}[t]\Large
	\begin{center}
		\scriptsize
		\setlength\tabcolsep{1.7pt}
		\renewcommand\arraystretch{1.6}
		\caption{Comparisons of different object detection methods on DOTA-v1.0 test set. The numbers in the brackets represent the gains of accuracy and speed in comparison with the baselines (the same below).}\label{table1}
		%\vspace{-3mm}
		\resizebox{1\textwidth}{!}{
			\begin{tabular}{lccccccccccccccccccc}
				\hline
				\multicolumn{1}{c|}{Method}                & \multicolumn{1}{c|}{Publication} & \multicolumn{1}{c|}{Backbone}  & \multicolumn{1}{c}{PL}    & \multicolumn{1}{c}{BD}    & \multicolumn{1}{c}{BR}    & \multicolumn{1}{c}{GTF}   & \multicolumn{1}{c}{SV}    & \multicolumn{1}{c}{LV}    & \multicolumn{1}{c}{SH}    & \multicolumn{1}{c}{TC}    & \multicolumn{1}{c}{BC}    & \multicolumn{1}{c}{ST}    & \multicolumn{1}{c}{SBF}   & \multicolumn{1}{c}{RA}    & \multicolumn{1}{c}{HA}    & \multicolumn{1}{c}{SP}    & \multicolumn{1}{c|}{HC}    & \multicolumn{1}{c}{mAP}          & FPS        \\ \hline
				\multicolumn{1}{c|}{\textbf{\emph{{One-stage}}}}    & \multicolumn{19}{l}{} 
				\\ \hline
				\multicolumn{1}{c|}{RetinaNet-O{\cite{retinanet_iccv17}}}           & \multicolumn{1}{c|}{ICCV2017}  & \multicolumn{1}{c|}{R50-FPN}   & \multicolumn{1}{c}{88.67} & \multicolumn{1}{c}{77.62} & \multicolumn{1}{c}{41.81} & \multicolumn{1}{c}{58.17} & \multicolumn{1}{c}{74.58} & \multicolumn{1}{c}{71.64} & \multicolumn{1}{c}{79.11} & \multicolumn{1}{c}{90.29} & \multicolumn{1}{c}{82.18} & \multicolumn{1}{c}{74.32} & \multicolumn{1}{c}{54.75} & \multicolumn{1}{c}{60.60} & \multicolumn{1}{c}{62.57} & \multicolumn{1}{c}{69.67} & \multicolumn{1}{c|}{60.64} & \multicolumn{1}{c}{68.43}        & 16.9       \\ %\hline
				
				\multicolumn{1}{c|}{DRN{\cite{drn_cvpr20}}}                   & \multicolumn{1}{c|}{CVPR2020}  & \multicolumn{1}{c|}{H-104}     & \multicolumn{1}{c}{88.91} & \multicolumn{1}{c}{80.22} & \multicolumn{1}{c}{43.52} & \multicolumn{1}{c}{63.35} & \multicolumn{1}{c}{73.48} & \multicolumn{1}{c}{70.69} & \multicolumn{1}{c}{84.94} & \multicolumn{1}{c}{90.14} & \multicolumn{1}{c}{83.85} & \multicolumn{1}{c}{84.11} & \multicolumn{1}{c}{50.12} & \multicolumn{1}{c}{58.41} & \multicolumn{1}{c}{67.62} & \multicolumn{1}{c}{68.60} & \multicolumn{1}{c|}{52.50} & \multicolumn{1}{c}{70.70}        & -          \\ %\hline
				
				\multicolumn{1}{c|}{PIoU{\cite{piou_eccv20}}}                  & \multicolumn{1}{c|}{ECCV2020}  & \multicolumn{1}{c|}{DLA-34}    & \multicolumn{1}{c}{80.90} & \multicolumn{1}{c}{69.70} & \multicolumn{1}{c}{24.10} & \multicolumn{1}{c}{60.20} & \multicolumn{1}{c}{38.30} & \multicolumn{1}{c}{64.40} & \multicolumn{1}{c}{64.80} & \multicolumn{1}{c}{90.90} & \multicolumn{1}{c}{77.20} & \multicolumn{1}{c}{70.40} & \multicolumn{1}{c}{46.50} & \multicolumn{1}{c}{37.10} & \multicolumn{1}{c}{57.10} & \multicolumn{1}{c}{61.90} & \multicolumn{1}{c|}{64.00} & \multicolumn{1}{c}{60.50}        & -          \\ %\hline
				
				\multicolumn{1}{c|}{DAL{\cite{dal_aaai21}}}                   & \multicolumn{1}{c|}{AAAI2021}  & \multicolumn{1}{c|}{R50-FPN}   & \multicolumn{1}{c}{88.68} & \multicolumn{1}{c}{76.55} & \multicolumn{1}{c}{45.08} & \multicolumn{1}{c}{66.80} & \multicolumn{1}{c}{67.00} & \multicolumn{1}{c}{76.76} & \multicolumn{1}{c}{79.74} & \multicolumn{1}{c}{90.84} & \multicolumn{1}{c}{79.54} & \multicolumn{1}{c}{78.45} & \multicolumn{1}{c}{57.71} & \multicolumn{1}{c}{62.27} & \multicolumn{1}{c}{69.05} & \multicolumn{1}{c}{73.14} & \multicolumn{1}{c|}{60.11} & \multicolumn{1}{c}{71.44}        & -          \\ %\hline
				
				\multicolumn{1}{c|}{S2ANet{\cite{s2anet_tgrs22}}}                & \multicolumn{1}{c|}{TGRS2021}  & \multicolumn{1}{c|}{R50-FPN}   & \multicolumn{1}{c}{89.11} & \multicolumn{1}{c}{82.84} & \multicolumn{1}{c}{48.37} & \multicolumn{1}{c}{71.11} & \multicolumn{1}{c}{78.11} & \multicolumn{1}{c}{78.39} & \multicolumn{1}{c}{87.25} & \multicolumn{1}{c}{90.83} & \multicolumn{1}{c}{84.90} & \multicolumn{1}{c}{85.64} & \multicolumn{1}{c}{60.36} & \multicolumn{1}{c}{62.60} & \multicolumn{1}{c}{65.26} & \multicolumn{1}{c}{69.13} & \multicolumn{1}{c|}{57.94} & \multicolumn{1}{c}{74.12}        & 16.4       \\ %\hline
				
				\multicolumn{1}{c|}{R3Det{\cite{r3det_aaai21}}}                 & \multicolumn{1}{c|}{AAAI2021}  & \multicolumn{1}{c|}{R101-FPN}  & \multicolumn{1}{c}{88.76} & \multicolumn{1}{c}{83.09} & \multicolumn{1}{c}{50.91} & \multicolumn{1}{c}{67.27} & \multicolumn{1}{c}{76.23} & \multicolumn{1}{c}{80.39} & \multicolumn{1}{c}{86.72} & \multicolumn{1}{c}{90.78} & \multicolumn{1}{c}{84.68} & \multicolumn{1}{c}{84.68} & \multicolumn{1}{c}{61.98} & \multicolumn{1}{c}{61.35} & \multicolumn{1}{c}{66.91} & \multicolumn{1}{c}{70.63} & \multicolumn{1}{c|}{53.94} & \multicolumn{1}{c}{73.74}        & -          \\ %\hline
				
				\multicolumn{1}{c|}{DCL{\cite{dcl_cvpr21}}}                   & \multicolumn{1}{c|}{CVPR2021}  & \multicolumn{1}{c|}{R152-FPN}  & \multicolumn{1}{c}{89.10} & \multicolumn{1}{c}{84.13} & \multicolumn{1}{c}{50.15} & \multicolumn{1}{c}{73.57} & \multicolumn{1}{c}{71.48} & \multicolumn{1}{c}{58.13} & \multicolumn{1}{c}{78.00} & \multicolumn{1}{c}{90.89} & \multicolumn{1}{c}{86.64} & \multicolumn{1}{c}{86.78} & \multicolumn{1}{c}{67.97} & \multicolumn{1}{c}{67.25} & \multicolumn{1}{c}{65.63} & \multicolumn{1}{c}{74.06} & \multicolumn{1}{c|}{67.05} & \multicolumn{1}{c}{74.06}        & -          \\ %\hline
				
				\multicolumn{1}{c|}{GWD{\cite{gwd_icml21}}}                   & \multicolumn{1}{c|}{ICML2021}  & \multicolumn{1}{c|}{R101-FPN}  & \multicolumn{1}{c}{89.59} & \multicolumn{1}{c}{81.18} & \multicolumn{1}{c}{52.89} & \multicolumn{1}{c}{70.37} & \multicolumn{1}{c}{77.73} & \multicolumn{1}{c}{82.42} & \multicolumn{1}{c}{86.99} & \multicolumn{1}{c}{89.31} & \multicolumn{1}{c}{83.06} & \multicolumn{1}{c}{85.97} & \multicolumn{1}{c}{64.07} & \multicolumn{1}{c}{65.14} & \multicolumn{1}{c}{68.05} & \multicolumn{1}{c}{70.95} & \multicolumn{1}{c|}{58.45} & \multicolumn{1}{c}{74.09}        & -          \\ \hline
				
				\rowcolor[HTML]{EFEFEF}
				\multicolumn{1}{c|}{\textbf{RetinaNet-O + OAN}}     & \multicolumn{1}{c|}{-}         & \multicolumn{1}{c|}{R50-FPN}   & \multicolumn{1}{c}{89.20} & \multicolumn{1}{c}{78.63} & \multicolumn{1}{c}{39.21} & \multicolumn{1}{c}{68.99} & \multicolumn{1}{c}{78.39} & \multicolumn{1}{c}{62.97} & \multicolumn{1}{c}{77.18} & \multicolumn{1}{c}{90.66} & \multicolumn{1}{c}{81.43} & \multicolumn{1}{c}{81.23} & \multicolumn{1}{c}{55.55} & \multicolumn{1}{c}{63.64} & \multicolumn{1}{c}{53.58} & \multicolumn{1}{c}{65.96} & \multicolumn{1}{c|}{51.09} & \multicolumn{1}{c}{69.18(+0.75)} & 22.1(\textcolor{blue}{\textbf{+5.2}}) \\ %\hline
				
				\rowcolor[HTML]{EFEFEF}
				\multicolumn{1}{c|}{\textbf{S2ANet + OAN}}          & \multicolumn{1}{c|}{-}         & \multicolumn{1}{c|}{R50-FPN}   & \multicolumn{1}{c}{88.84} & \multicolumn{1}{c}{80.75} & \multicolumn{1}{c}{50.34} & \multicolumn{1}{c}{70.62} & \multicolumn{1}{c}{78.42} & \multicolumn{1}{c}{77.79} & \multicolumn{1}{c}{87.17} & \multicolumn{1}{c}{90.83} & \multicolumn{1}{c}{84.65} & \multicolumn{1}{c}{85.53} & \multicolumn{1}{c}{59.77} & \multicolumn{1}{c}{63.78} & \multicolumn{1}{c}{66.57} & \multicolumn{1}{c}{68.88} & \multicolumn{1}{c|}{61.62} & \multicolumn{1}{c}{74.37(+0.25)} & 21.8(\textcolor{blue}{\textbf{+5.4}}) \\ \hline
				\multicolumn{1}{c|}{\textbf{\emph{Two-stage}}}    & \multicolumn{19}{l}{}                                                                                                                                                                                                                                                                                                                                                                                                                                                                                                                                                                                 \\ \hline
				
				\multicolumn{1}{c|}{Faster R-CNN-O{\cite{faster_tpami17}}}        & \multicolumn{1}{c|}{TPAMI2017}  & \multicolumn{1}{c|}{R50-FPN}   & \multicolumn{1}{c}{88.44} & \multicolumn{1}{c}{73.06} & \multicolumn{1}{c}{44.86} & \multicolumn{1}{c}{59.09} & \multicolumn{1}{c}{73.25} & \multicolumn{1}{c}{71.49} & \multicolumn{1}{c}{77.11} & \multicolumn{1}{c}{90.84} & \multicolumn{1}{c}{78.94} & \multicolumn{1}{c}{83.90} & \multicolumn{1}{c}{48.59} & \multicolumn{1}{c}{62.95} & \multicolumn{1}{c}{62.18} & \multicolumn{1}{c}{64.91} & \multicolumn{1}{c|}{56.18} & \multicolumn{1}{c}{69.05}        & 14.1       \\ %\hline
				
				\multicolumn{1}{c|}{RoI Transformer{\cite{roitrans_cvpr19}}}       & \multicolumn{1}{c|}{CVPR2019}  & \multicolumn{1}{c|}{R50-FPN}   & \multicolumn{1}{c}{88.34} & \multicolumn{1}{c}{77.07} & \multicolumn{1}{c}{51.63} & \multicolumn{1}{c}{69.62} & \multicolumn{1}{c}{77.45} & \multicolumn{1}{c}{77.15} & \multicolumn{1}{c}{87.11} & \multicolumn{1}{c}{90.75} & \multicolumn{1}{c}{84.90} & \multicolumn{1}{c}{83.14} & \multicolumn{1}{c}{52.95} & \multicolumn{1}{c}{63.75} & \multicolumn{1}{c}{74.45} & \multicolumn{1}{c}{74.45} & \multicolumn{1}{c|}{59.24} & \multicolumn{1}{c}{73.76}        & 12.4       \\ %\hline
				
				\multicolumn{1}{c|}{SCRDet{\cite{scrdet_iccv19}}}                & \multicolumn{1}{c|}{ICCV2019}  & \multicolumn{1}{c|}{R-101-FPN} & \multicolumn{1}{c}{89.98} & \multicolumn{1}{c}{80.65} & \multicolumn{1}{c}{52.09} & \multicolumn{1}{c}{68.36} & \multicolumn{1}{c}{68.36} & \multicolumn{1}{c}{60.32} & \multicolumn{1}{c}{72.41} & \multicolumn{1}{c}{90.85} & \multicolumn{1}{c}{87.94} & \multicolumn{1}{c}{86.86} & \multicolumn{1}{c}{65.02} & \multicolumn{1}{c}{66.68} & \multicolumn{1}{c}{66.25} & \multicolumn{1}{c}{68.24} & \multicolumn{1}{c|}{65.21} & \multicolumn{1}{c}{72.61}        & -          \\ %\hline
				
				\multicolumn{1}{c|}{Gliding vertex{\cite{gv_tpami21}}}        & \multicolumn{1}{c|}{TPAMI2020} & \multicolumn{1}{c|}{R-101-FPN} & \multicolumn{1}{c}{89.64} & \multicolumn{1}{c}{85.00} & \multicolumn{1}{c}{52.26} & \multicolumn{1}{c}{77.34} & \multicolumn{1}{c}{73.01} & \multicolumn{1}{c}{73.14} & \multicolumn{1}{c}{86.82} & \multicolumn{1}{c}{90.74} & \multicolumn{1}{c}{79.02} & \multicolumn{1}{c}{86.81} & \multicolumn{1}{c}{59.55} & \multicolumn{1}{c}{70.91} & \multicolumn{1}{c}{72.94} & \multicolumn{1}{c}{70.86} & \multicolumn{1}{c|}{57.32} & \multicolumn{1}{c}{75.01}        & 13.2       \\ %\hline
				
				\multicolumn{1}{c|}{CFA{\cite{cfa_cvpr21}}}                   & \multicolumn{1}{c|}{CVPR2021}  & \multicolumn{1}{c|}{R101-FPN}  & \multicolumn{1}{c}{89.26} & \multicolumn{1}{c}{81.72} & \multicolumn{1}{c}{51.81} & \multicolumn{1}{c}{67.17} & \multicolumn{1}{c}{79.99} & \multicolumn{1}{c}{78.25} & \multicolumn{1}{c}{84.46} & \multicolumn{1}{c}{90.77} & \multicolumn{1}{c}{83.40} & \multicolumn{1}{c}{85.54} & \multicolumn{1}{c}{54.86} & \multicolumn{1}{c}{67.75} & \multicolumn{1}{c}{73.04} & \multicolumn{1}{c}{70.24} & \multicolumn{1}{c|}{64.96} & \multicolumn{1}{c}{75.05}        & -          \\ %\hline
				
				\multicolumn{1}{c|}{ReDet{\cite{redet_cvpr21}}}                 & \multicolumn{1}{c|}{CVPR2021}  & \multicolumn{1}{c|}{Re50-FPN}  & \multicolumn{1}{c}{88.79} & \multicolumn{1}{c}{82.64} & \multicolumn{1}{c}{53.97} & \multicolumn{1}{c}{74.00} & \multicolumn{1}{c}{78.13} & \multicolumn{1}{c}{84.06} & \multicolumn{1}{c}{88.04} & \multicolumn{1}{c}{90.89} & \multicolumn{1}{c}{87.78} & \multicolumn{1}{c}{85.75} & \multicolumn{1}{c}{61.76} & \multicolumn{1}{c}{60.39} & \multicolumn{1}{c}{75.96} & \multicolumn{1}{c}{68.07} & \multicolumn{1}{c|}{63.59} & \multicolumn{1}{c}{76.25}        & 3.1        \\ %\hline
				
				\multicolumn{1}{c|}{Oriented R-CNN{\cite{orcnn_iccv21}}}       & \multicolumn{1}{c|}{ICCV2021}  & \multicolumn{1}{c|}{R50-FPN}   & \multicolumn{1}{c}{89.46} & \multicolumn{1}{c}{82.12} & \multicolumn{1}{c}{54.78} & \multicolumn{1}{c}{70.86} & \multicolumn{1}{c}{78.93} & \multicolumn{1}{c}{83.00} & \multicolumn{1}{c}{88.20} & \multicolumn{1}{c}{90.90} & \multicolumn{1}{c}{87.50} & \multicolumn{1}{c}{84.68} & \multicolumn{1}{c}{63.97} & \multicolumn{1}{c}{67.69} & \multicolumn{1}{c}{74.94} & \multicolumn{1}{c}{68.84} & \multicolumn{1}{c|}{52.28} & \multicolumn{1}{c}{75.87}        & 15.6       \\ %\hline
				
				\multicolumn{1}{c|}{Oriented R-CNN{\cite{orcnn_iccv21}}}        & \multicolumn{1}{c|}{ICCV2021}  & \multicolumn{1}{c|}{RXt50-FPN} & \multicolumn{1}{c}{89.76} & \multicolumn{1}{c}{83.97} & \multicolumn{1}{c}{55.21} & \multicolumn{1}{c}{74.80} & \multicolumn{1}{c}{78.60} & \multicolumn{1}{c}{83.24} & \multicolumn{1}{c}{88.28} & \multicolumn{1}{c}{90.91} & \multicolumn{1}{c}{86.58} & \multicolumn{1}{c}{85.62} & \multicolumn{1}{c}{62.69} & \multicolumn{1}{c}{60.98} & \multicolumn{1}{c}{75.26} & \multicolumn{1}{c}{69.67} & \multicolumn{1}{c|}{62.57} & \multicolumn{1}{c}{76.54}        & 13.4       \\ \hline
				
				\rowcolor[HTML]{EFEFEF}
				\multicolumn{1}{c|}{\textbf{Faster R-CNN-O + OAN}}  & \multicolumn{1}{c|}{-}         & \multicolumn{1}{c|}{R50-FPN}   & \multicolumn{1}{c}{88.44} & \multicolumn{1}{c}{76.33} & \multicolumn{1}{c}{46.31} & \multicolumn{1}{c}{59.70} & \multicolumn{1}{c}{73.30} & \multicolumn{1}{c}{72.13} & \multicolumn{1}{c}{77.90} & \multicolumn{1}{c}{90.72} & \multicolumn{1}{c}{79.02} & \multicolumn{1}{c}{81.60} & \multicolumn{1}{c}{44.80} & \multicolumn{1}{c}{58.66} & \multicolumn{1}{c}{61.28} & \multicolumn{1}{c}{67.51} & \multicolumn{1}{c|}{62.87} & \multicolumn{1}{c}{69.37(+0.32)} & 18.5(\textcolor{blue}{\textbf{+4.4}}) \\ %\hline
				\rowcolor[HTML]{EFEFEF}
				\multicolumn{1}{c|}{\textbf{RoI Transformer + OAN}} & \multicolumn{1}{c|}{-}         & \multicolumn{1}{c|}{R50-FPN}   & \multicolumn{1}{c}{88.46} & \multicolumn{1}{c}{78.22} & \multicolumn{1}{c}{52.24} & \multicolumn{1}{c}{67.34} & \multicolumn{1}{c}{78.11} & \multicolumn{1}{c}{76.72} & \multicolumn{1}{c}{86.92} & \multicolumn{1}{c}{90.52} & \multicolumn{1}{c}{86.26} & \multicolumn{1}{c}{77.06} & \multicolumn{1}{c}{57.93} & \multicolumn{1}{c}{63.40} & \multicolumn{1}{c}{73.87} & \multicolumn{1}{c}{69.53} & \multicolumn{1}{c|}{62.21} & \multicolumn{1}{c}{73.92(+0.16)} & 16.6(\textcolor{blue}{\textbf{+4.2}}) \\ %\hline
				\rowcolor[HTML]{EFEFEF}
				\multicolumn{1}{c|}{\textbf{Oriented R-CNN + OAN}}  & \multicolumn{1}{c|}{-}         & \multicolumn{1}{c|}{R50-FPN}   & \multicolumn{1}{c}{89.37} & \multicolumn{1}{c}{82.78} & \multicolumn{1}{c}{54.33} & \multicolumn{1}{c}{71.80} & \multicolumn{1}{c}{78.93} & \multicolumn{1}{c}{83.03} & \multicolumn{1}{c}{88.22} & \multicolumn{1}{c}{90.90} & \multicolumn{1}{c}{87.57} & \multicolumn{1}{c}{84.75} & \multicolumn{1}{c}{62.65} & \multicolumn{1}{c}{65.61} & \multicolumn{1}{c}{74.35} & \multicolumn{1}{c}{69.42} & \multicolumn{1}{c|}{56.57} & \multicolumn{1}{c}{76.02(+0.15)} & 20.4(\textcolor{blue}{\textbf{+4.8}}) \\ %\hline
				\rowcolor[HTML]{EFEFEF}
				\multicolumn{1}{c|}{\textbf{Oriented R-CNN + OAN}}  & \multicolumn{1}{c|}{-}         & \multicolumn{1}{c|}{RXt50-FPN}   & \multicolumn{1}{c}{89.70} & \multicolumn{1}{c}{84.03} & \multicolumn{1}{c}{54.61} & \multicolumn{1}{c}{73.46} & \multicolumn{1}{c}{79.30} & \multicolumn{1}{c}{83.27} & \multicolumn{1}{c}{88.12} & \multicolumn{1}{c}{90.90} & \multicolumn{1}{c}{85.22} & \multicolumn{1}{c}{84.68} & \multicolumn{1}{c}{62.06} & \multicolumn{1}{c}{66.87} & \multicolumn{1}{c}{75.28} & \multicolumn{1}{c}{70.71} & \multicolumn{1}{c|}{62.75} & \multicolumn{1}{c}{76.73(+0.19)} & 17.5(\textcolor{blue}{\textbf{+4.1}}) \\ 
				\hline
		\end{tabular}}
	\end{center}
	\vspace{-1mm}
\end{table*}

\begin{table*}[h]\Large
	\begin{center}
		\scriptsize
		\setlength\tabcolsep{1.5pt}
		\renewcommand\arraystretch{1.6}
		\caption{Comparisons of four baseline methods and their combinations with our OAN on DOTA-v1.5 test set.}\label{table2}
		%\vspace{-3mm}
		\resizebox{1\textwidth}{!}{
			\begin{tabular}{c|c|cccccccccccccccc|cc}
				\hline
				{Method}                & \multicolumn{1}{c|}{Backbone} & \multicolumn{1}{c}{PL} & \multicolumn{1}{c}{BD} & \multicolumn{1}{c}{BR} & \multicolumn{1}{c}{GTF} & \multicolumn{1}{c}{SV} & \multicolumn{1}{c}{LV} & \multicolumn{1}{c}{SH} & \multicolumn{1}{c}{TC} & \multicolumn{1}{c}{BC} & \multicolumn{1}{c}{ST} & \multicolumn{1}{c}{SBF} & \multicolumn{1}{c}{RA} & \multicolumn{1}{c}{HA} & \multicolumn{1}{c}{SP} & \multicolumn{1}{c}{HC} & \multicolumn{1}{c|}{CC} & mAP          & FPS                         \\ \hline
				RetinaNet-O{\cite{retinanet_iccv17}}           & \multicolumn{1}{c|}{R50-FPN}  & 71.43                   & 77.64                   & 42.12                   & 64.65                    & 44.53                   & 56.79                   & 73.31                   & 90.84                   & 76.02                   & 59.96                   & 46.95                    & 69.24                   & 59.65                   & 64.52                   & 48.06                   & 0.83                    & 59.16        & 16.9                        \\ %\hline
				Faster R-CNN-O{\cite{faster_tpami17}}        & R50-FPN                       & 71.89                   & 74.47                   & 44.45                   & 59.87                    & 51.28                   & 68.98                   & 79.37                   & 90.78                   & 77.38                   & 67.50                   & 47.75                    & 69.72                   & 61.22                   & 65.28                   & 60.47                   & 1.54                    & 62.00        & 14.1                        \\ %\hline
				RoI Transformer{\cite{roitrans_cvpr19}}       & R50-FPN                       & 71.92                   & 76.07                   & 51.87                   & 69.24                    & 52.05                   & 75.18                   & 80.72                   & 90.53                   & 78.58                   & 68.26                   & 49.18                    & 71.74                   & 67.51                   & 65.53                   & 62.16                   & 9.99                    & 65.03        & 12.4                        \\ %\hline
				Oriented R-CNN{\cite{orcnn_iccv21}}        & R50-FPN                       & 79.95                   & 81.00                   & 53.90                   & 70.59                    & 52.48                   & 76.21                   & 86.98                   & 90.88                   & 78.33                   & 68.26                   & 58.94                    & 72.60                   & 72.75                   & 65.32                   & 58.18                   & 3.72                    & 66.88        & 15.6                        \\ \hline
				
				\rowcolor[HTML]{EFEFEF}
				\textbf{RetinaNet-O + OAN}     & R50-FPN                       & 76.19                   & 78.71                   & 40.52                   & 65.56                    & 49.08                   & 59.84                   & 77.55                   & 90.85                   & 75.10                   & 69.57                   & 44.74                    & 69.58                   & 54.57                   & 63.20                   & 41.36                   & 1.07                    & 59.84(+0.68) & 22.1(\textcolor{blue}{\textbf{+5.2}})                        \\ %\hline
				
				\rowcolor[HTML]{EFEFEF}
				\textbf{Faster R-CNN-O + OAN}  & R50-FPN                       & 72.10                   & 75.56                   & 46.18                   & 62.91                    & 51.51                   & 69.86                   & 79.93                   & 90.87                   & 78.33                   & 67.93                   & 52.37                    & 68.60                   & 62.25                   & 65.81                   & 61.44                   & 1.07                    & 62.92(+0.92) & 18.5(\textcolor{blue}{\textbf{+4.4}})                        \\ %\hline
				
				\rowcolor[HTML]{EFEFEF}
				\textbf{RoI Transformer + OAN} & R50-FPN                       & 72.27                   & 77.75                   & 47.75                   & 68.12                    & 52.35                   & 76.04                   & 80.95                   & 90.88                   & 79.36                   & 68.69                   & 60.31                    & 65.52                   & 68.32                   & 66.86                   & 63.48                   & 12.72                   & 65.71(+0.68) & 16.6(\textcolor{blue}{\textbf{+4.2}}) \\ %\hline
				
				\rowcolor[HTML]{EFEFEF}
				\textbf{Oriented R-CNN + OAN}  & R50-FPN                       & 79.99                   & 80.29                   & 53.33                   & 71.83                    & 52.52                   & 76.91                   & 87.73                   & 90.83                   & 81.59                   & 68.74                   & 58.11                    & 72.98                   & 67.68                   & 64.93                   & 54.34                   & 9.31                    & 66.95(+0.07) & 20.4(\textcolor{blue}{\textbf{+4.8}})                        \\ \hline
				
		\end{tabular}}
	\end{center}
	
\end{table*}

\begin{table*}[h]
	\begin{center}
		\scriptsize
		\setlength\tabcolsep{1.5pt}
		\renewcommand\arraystretch{1.6}
		\caption{Comparisons of four baseline methods and their combinations with our OAN on DOTA-v2.0 test set.}\label{table3}
		%\vspace{-3mm}
		\resizebox{1\textwidth}{!}{
			\begin{tabular}{c|c|cccccccccccccccccc|cc}
				\hline
				{Method}                & \multicolumn{1}{c|}{Backbone} & \multicolumn{1}{c}{PL} & \multicolumn{1}{c}{BD} & \multicolumn{1}{c}{BR} & \multicolumn{1}{c}{GTF} & \multicolumn{1}{c}{SV} & \multicolumn{1}{c}{LV} & \multicolumn{1}{c}{SH} & \multicolumn{1}{c}{TC} & \multicolumn{1}{c}{BC} & \multicolumn{1}{c}{ST} & \multicolumn{1}{c}{SBF} & \multicolumn{1}{c}{RA} & \multicolumn{1}{c}{HA} & \multicolumn{1}{c}{SP} & \multicolumn{1}{c}{HC} & \multicolumn{1}{c}{CC} & Air   & Heli  & \multicolumn{1}{c}{mAP} & \multicolumn{1}{c}{FPS}    \\ \hline
				RetinaNet-O{\cite{retinanet_iccv17}}           & \multicolumn{1}{c|}{R50-FPN}  & 70.63                   & 47.26                   & 39.12                   & 55.02                    & 38.10                    & 40.52                   & 47.16                   & 77.74                   & 56.86                   & 52.12                   & 37.22                    & 51.75                   & 44.15                   & 53.19                   & 51.06                   & 6.58                    & 64.28 & 7.45  & 46.68                    & 16.9                        \\ %\hline
				Faster R-CNN-O{\cite{faster_tpami17}}        & R50-FPN                       & 71.61                   & 47.20                    & 39.28                   & 58.70                     & 35.55                   & 48.88                   & 51.51                   & 78.97                   & 58.36                   & 58.55                   & 36.11                    & 51.73                   & 43.57                   & 55.33                   & 57.07                   & 3.51                    & 52.94 & 2.79  & 47.31                    & 14.1                        \\ %\hline
				RoI Transformer{\cite{roitrans_cvpr19}}       & R50-FPN                       & 71.81                   & 48.39                   & 45.88                   & 64.02                    & 42.09                   & 54.39                   & 59.92                   & 82.70                    & 63.29                   & 58.71                   & 41.04                    & 52.82                   & 53.32                   & 56.18                   & 57.94                   & 25.71                   & 63.72 & 8.70   & 52.81                    & 12.4                        \\ %\hline
				Oriented R-CNN{\cite{orcnn_iccv21}}        & R50-FPN                       & 78.65                   & 51.80                   & 47.15                   & 65.78                    & 43.35                   & 58.29                   & 60.89                   & 82.83                   & 63.51                   & 59.50                   & 43.40                    & 55.79                   & 52.90                   & 56.18                   & 54.13                   & 27.55                   & 66.24 & 5.22  & 54.06                    & 15.6                        \\ \hline
				\rowcolor[HTML]{EFEFEF}
				\textbf{RetinaNet-O + OAN}     & R50-FPN                       & 73.93                   & 52.03                   & 38.23                   & 54.06                    & 44.11                   & 45.20                   & 51.63                   & 78.28                   & 61.19                   & 59.64                   & 38.82                    & 50.54                   & 43.19                   & 54.32                   & 45.96                   & 10.81                   & 43.96 & 0.72  & 47.03(+0.35)             & 22.1(\textcolor{blue}{\textbf{+5.2}})                        \\ %\hline
				\rowcolor[HTML]{EFEFEF}
				\textbf{Faster R-CNN-O + OAN}  & R50-FPN                       & 71.85                   & 47.86                   & 40.51                   & 59.50                    & 35.64                   & 50.44                   & 51.93                   & 78.44                   & 59.42                   & 59.56                   & 39.45                    & 52.39                   & 43.99                   & 56.33                   & 53.64                   & 5.81                    & 47.19 & 12.38 & 48.13(+0.82)             & 18.5(\textcolor{blue}{\textbf{+4.4}})                        \\ %\hline
				\rowcolor[HTML]{EFEFEF}
				\textbf{RoI Transformer + OAN} & R50-FPN                       & 71.90                   & 50.48                   & 43.55                   & 65.76                    & 42.51                   & 54.90                   & 60.35                   & 79.09                   & 63.45                   & 60.03                   & 46.59                    & 52.91                   & 54.30                   & 57.25                   & 60.33                   & 23.59                   & 66.39 & 9.72  & 53.51(+0.70)              & 16.6(\textcolor{blue}{\textbf{+4.2}}) \\ %\hline
				\rowcolor[HTML]{EFEFEF}
				\textbf{Oriented R-CNN + OAN}  & R50-FPN                       & 79.18                   & 51.57                   & 47.50                   & 66.61                    & 43.30                   & 58.07                   & 60.73                   & 82.85                   & 64.47                   & 59.62                   & 44.31                    & 56.66                   & 52.71                   & 56.73                   & 53.04                   & 26.10                   & 66.41 & 14.42 & 54.68(+0.62)             & 20.4(\textcolor{blue}{\textbf{+4.8}})                        \\ \hline
		\end{tabular}}
	\end{center}
	\vspace{-1mm}
\end{table*}
\subsubsection{Results on DOTA-v1.0}
Table \ref{table1} reports the results on DOTA-v1.0 test set. As shown, our approach outperforms all five baselines in terms of both accuracy and speed. Especially, with the help of OAN, all five detectors acquire more than 30.0\% speed-up. To be specific, when applying OAN to one-stage detectors: (i) RetinaNet-O with OAN runs at 22.1 FPS, which improves the detection speed by a big margin of 30.7\%, while boosting the detection accuracy with 0.75\% mAP. (ii) S2ANet with OAN obtains the best accuracy (74.37\% mAP) while with competitive speed (21.8 FPS) among all recent one-stage detectors.
For two-stage detectors, our OAN boosts the detection speeds of Faster R-CNN-O, RoI Transformer, and Oriented R-CNN by 31.2\%, 33.8\%, and 30.7\%, respectively, meanwhile improving the accuracy to some extent. Based on RXt50-FPN, Oriented R-CNN with OAN achieves the highest accuracy (76.73\% mAP), surpassing all one-stage and two-stage detectors. Compared with the strongest competitor ReDet in accuracy, Oriented R-CNN with OAN is more than six times faster. Note that ReDet uses specific-purpose rotation-equivariant backbone, which is much slower than others. From these results, we concluded that our OAN leads to significant speed gains (more than 30.0\%) while achieving consistent improvements of accuracy for all five detectors.

\subsubsection{Results on DOTA-v1.5 and DOTA-v2.0}	
Table \ref{table2} and Table \ref{table3} give the comparisons of four baseline methods (RetinaNet-O, Faster R-CNN-O, RoI Transformer and Oriented R-CNN) and their combinations with our OAN on the DOTA-v1.5 and DOTA-v2.0 datasets, respectively. For fair comparisons, we use R50-FPN as the backbone. Following common practice{\cite{dota20_tpami21}}, RetinaNet-O is trained with 2$\times$ learning schedule (24 epochs). As shown, similar to the results of DOTA-v1.0, all four baselines with OAN obtain consistent improvements of speed (30.7\%, 31.2\%, 33.8\%, and 30.7\% speed gains, respectively) while surpassing the detection accuracies of all baselines. Running at 20.4 FPS, the strong baseline Oriented R-CNN with OAN can achieve 66.95\% mAP and 54.68\% mAP on DOTA-v1.5 and DOTA-v2.0 datasets, respectively. Both of them are the highest accuracies on these two datasets.

\subsubsection{Results on Extremely Large Images}

In order to be closer to real-world applications, we use all 10 Gaofen-2 satellite images (29200$\times$27620 pixels), which are extremely large compared with natural images, from DOTA-v2.0 test-dev for further experiments. Here, each Gaofen-2 image is divided into 1224 patches with the size of 1024$\times$1024, thus obtaining a total of 12240 patches (this number exceeds that of test set of DOTA-v1.0).
Here, we train our models using R50-FPN on DOTA-v2.0 and only submit the results of 10 Gaofen-2 images to evaluation 

\begin{figure*}[h]
	\begin{center}
		\includegraphics[width=0.98\linewidth]{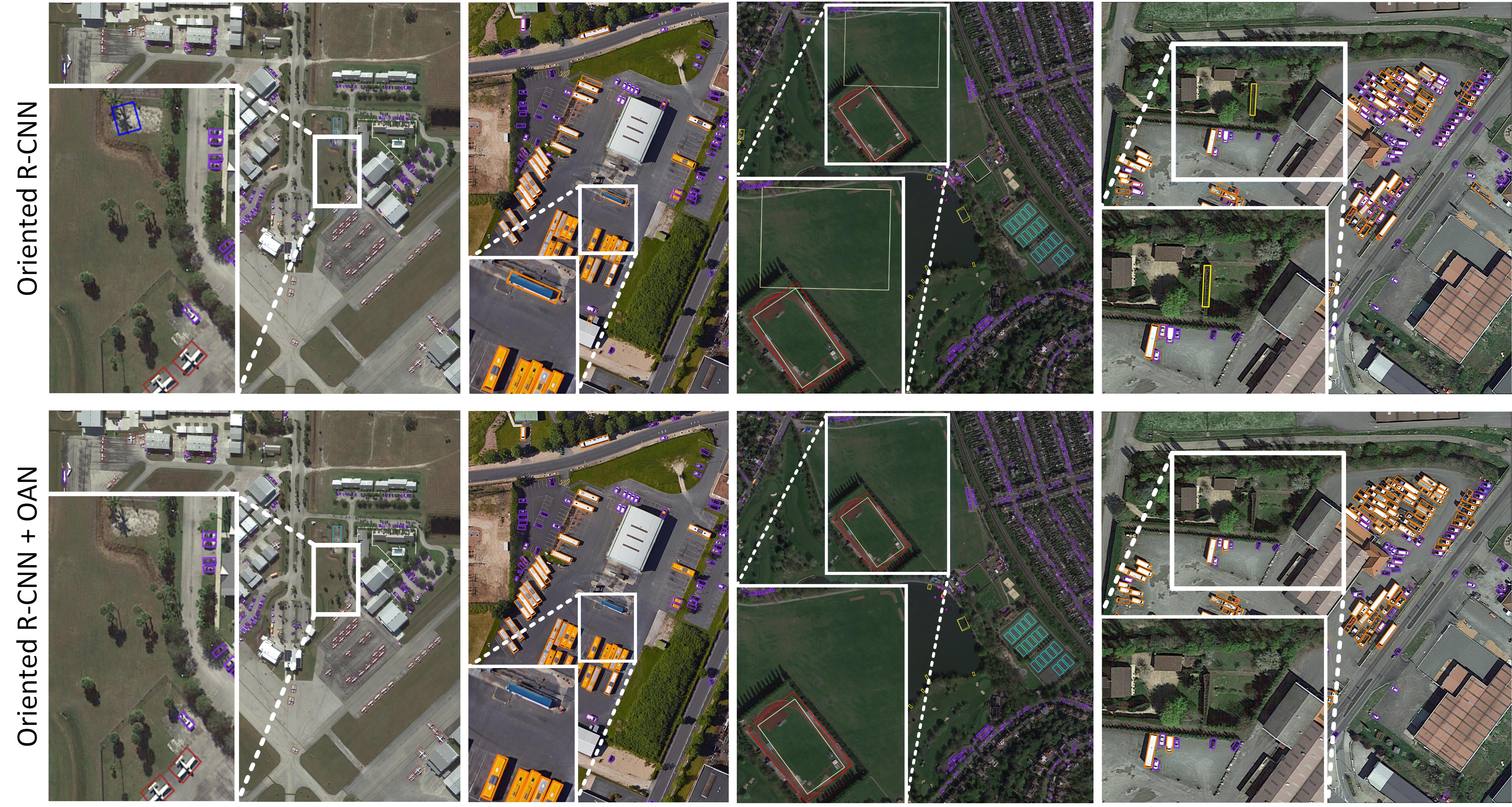}\\
	\end{center}
	\vspace*{-5mm}
	\caption{Detection results of Oriented R-CNN and Oriented R-CNN with OAN. Our OAN can help detector reduce false alarms. Here, each color represents one object class (the same below). It is worthy noting that, for better comparison, the results are cropped from the original large images.}\label{Fig7}
	\vspace{1mm}
\end{figure*}	

\begin{figure*}[h]
\hspace{3mm}
\begin{minipage}[h]{0.45\textwidth}
	%\centering
	\makeatletter\def\@captype{table}
	\scriptsize
	\setlength\tabcolsep{1pt}
	\renewcommand\arraystretch{1.4}
	\caption{Detection results on 10 Gaofen-2 images.}\label{table4}
	\vspace{-2mm}
	\resizebox{1\textwidth}{!}
	{
		\begin{tabular}{c|c|c}
			\hline
			Method                                                               & mAP          & FPS         \\ \hline
			Oriented R-CNN\cite{orcnn_iccv21}    & 10.26        & 15.6       \\ \hline
			\rowcolor[HTML]{EFEFEF}    
			\textbf{Oriented R-CNN + OAN}                                                       & 10.70(+0.44) & 26.6(\textcolor{blue}{\textbf{+11.0}}) \\ \hline
	\end{tabular}}
\end{minipage}
\hspace{3mm}
\begin{minipage}[h]{0.45\textwidth}
	\centering
	\makeatletter\def\@captype{table}
	\scriptsize
	\setlength\tabcolsep{0.8pt}
	\renewcommand\arraystretch{1.45}
	\caption{Benefit of OAN for detector training.}\label{table5}
	\vspace{-2mm}
	\resizebox{1\textwidth}{!}
	{
		\begin{tabular}{c|c}
			\hline
			Method                                   & mAP   \\ \hline
			Oriented R-CNN\cite{orcnn_iccv21}                           & 75.87 \\ \hline
			\rowcolor[HTML]{EFEFEF}
			\textbf{Oriented R-CNN + OAN (Only for training)} & \textbf{76.20} \\ \hline
	\end{tabular}}
\end{minipage}
\vspace{2mm}
\end{figure*}

\noindent server for obtaining the mAP results (the ground truth labels are not available). It is worth noting that the mAP values are much lower than Table \ref{table3} because the mAP is calculated only according to the outputs of those 10 images. Thus, they can not reflect the real accuracy. Table \ref{table4} presents the comparison results of Oriented R-CNN and Oriented R-CNN with OAN. As presented, our OAN increases the inference speed of the strong baseline Oriented R-CNN by 70.5\%, from 15.6 FPS to 26.6 FPS, but without sacrificing the accuracy. More specifically, 
given a Gaofen-2 satellite image with 29200$\times$27620 pixels, Oriented R-CNN takes 80 seconds for detection on a Tesla V100 GPU, and meanwhile, as a comparison, Oriented R-CNN with OAN only spends 46 seconds. 

\subsection{Analysis and Discussion} \label{section5-4}
According to the above experiments, we conclude that OAN can lead to noticeable speed improvements for baselines while boosting the accuracy to some extent. This looks harmonious, but the deep understanding of this phenomenon is still compelling, such as why OAN can improve detection speed and accuracy? Why there exists a big gap between DOTA datasets and Gaofen-2 images in terms of speed improvement?

To push the envelope further, we deeply analyze the experiments and identify the central causes. Obviously, the reason for the speed improvement comes from the reduction of patches participating in the detectors. Regarding the accuracy improvement, we attribute it to the benefits of multi-task training and the reduction of false alarms. As shown in Table \ref{table5}, adding OAN to Oriented R-CNN for training only (using all patches for testing without OAN) can lead to 0.33 points gain of mAP (75.87\% \emph{vs.} 76.20\%). Meanwhile, after eliminating \emph{invalid patches} by OAN, we can also reduce some false alarms, as shown in Figure \ref{Fig7}.
As for the big gap in speed gain between DOTA datasets and Gaofen-2 images, it is decided by the sparsity degree of the objects in aerial images. As pointed out by {\cite{dota20_tpami21}}, most images of DOTA-v1.0, DOTA-v1.5 and DOTA-v2.0 are selected areas to include many object instances from large-size images, so the number of \emph{invalid patches} is relatively less (about 39\% in train-val set of DOTA-v1.0). Thus, the improvement of speed on these three datasets is limited. Whereas the extremely large images collected from the Gaofen-2 Satellite have lower foreground ratios and much approach to the object distribution in real-world applications, so the \emph{invalid patches} account for a higher proportion (more than 75\%). Therefore, we achieve considerable improvement of detection speed in large aerial images.

\subsection{Ablation Studies}

OAN is the core for speeding up the detection in large aerial images. It tells detectors to take which patches as input and abandon the remaining ones. To study the effectiveness of OAN, we perform a number of ablation studies. We first investigate the behavior of OAN design, including feature map selection, the division of grids, sample assignment, and analyzing why OAN works. Then, we compare OAN to its most related competitors, including  $R^2$-CNN\cite{r2cnn_tgrs19}, ClusDet\cite{clusdet_iccv19}, and DMNet\cite{dmnet_cvprw20}), in terms of accuracy and speed. Next, we provide the analysis between speed and accuracy for OAN. Finally, We validate the robustness of OAN to different backbones. If not otherwise specified, for all experiments, we use Oriented R-CNN with R50-FPN as the baseline and report the results on DOTA-v1.0 dataset. 

\begin{figure*}
	\begin{center}
		\includegraphics[width=0.98\linewidth]{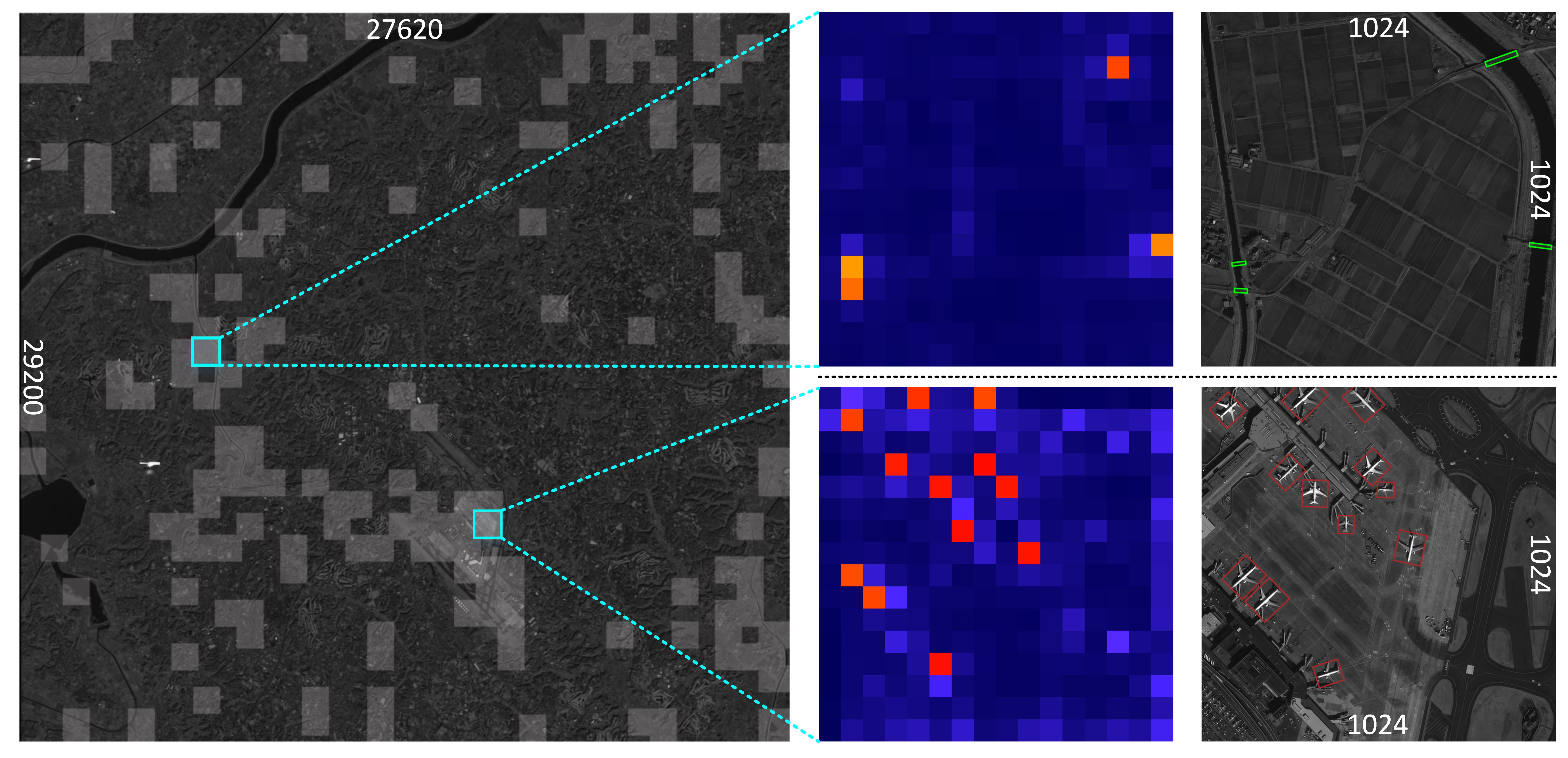}\\
	\end{center}
	\vspace*{-5mm}
	\caption{172 activated patches (1024$\times$1024) on a Gaofen-2 image (29200$\times$27620) and two zoomed-in patches with the objectness activation maps and corresponding detection results. It is worthy noting that, using our OAN, only 14\% of the total 1224 patches are activated.}\label{Fig8}
	%\vspace{-5mm}
\end{figure*}
\subsubsection{The Design of OAN}

\textbf{Which feature map is most suitable?} In theory, OAN can take the feature maps from any stages of backbones as input. For simplicity, the features from 2nd to 5th stages of ResNet50 are denoted as \{C2, C3, C4, C5\}.  When using C4 as input, we first apply a 3$\times$3 stride-2 convolutional layer to obtain the lower-dimensional features with the size of 32$\times$32$\times$256, and then reshape them into 16$\times$16$\times$1024 features for inference. In the same way, we can also take C2  or C3  as the input of OAN. In Table \ref{table6}, we show the comparison results of OAN by utilizing the features of different stages. Oriented R-CNN with OAN obtains the best accuracy when taking C5 as input. We attribute it to the stronger semantic representation of C5 since the feature maps with richer semantics are conducive to objectness prediction.     

\textbf{How important is the grid size?} The size of grids is one of the important factors for OAN. Here, we use C5 as the input of OAN and the size of the image patches, cropped from the original big images, is 1024$\times$1024. To adapt the division of different grid sizes, we conduct operations on the features (with the size of 16$\times$16$\times$256) produced by attaching a 3$\times$3 stride-2 convolutional layer to C5. For instance, when dividing the input patch into 8$\times$8 grids, the grid size is 128$\times$128 and the prediction output is 8$\times$8 

\hspace{-2mm}
\begin{minipage}[h]{0.47\textwidth}
	\centering
	\makeatletter\def\@captype{table}
	\scriptsize
	\setlength\tabcolsep{12pt}
	\renewcommand\arraystretch{1.2}
	\caption{Ablation studies of feature map selection.}\label{table6}
	\vspace{-2mm}
	\resizebox{1\textwidth}{!}
	{
		\begin{tabular}{c|c|c}  %|c|c|c|
			\hline
			Method                                & Feature & mAP   \\ \hline
			\multirow{4}{*}{Oriented R-CNN + OAN} & C2      & 63.63 \\ %\cline{2-3} 
			& C3      & 66.77 \\ %\cline{2-3} 
			& C4      & 72.06 \\ %\cline{2-3} 
			& \cellcolor[HTML]{EFEFEF}\textbf{C5}      & \cellcolor[HTML]{EFEFEF}\textbf{76.02} \\ \hline
	\end{tabular}}
	\vspace{1.2mm}
	\makeatletter\def\@captype{table}
	\scriptsize
	\setlength\tabcolsep{12pt}
	\renewcommand\arraystretch{1.4}
	\caption{Ablation studies of grid size.}\label{table7}
	\vspace{1mm}
	\resizebox{1\textwidth}{!}
	{
		\begin{tabular}{c|c|c}  %|c|c|c|
			\hline
			Method                                & Grid Size & mAP                        \\ \hline
			\multirow{6}{*}{Oriented R-CNN + OAN} & 1024$\times$1024    & 71.40                      \\ %\cline{2-3} 
			& 512$\times$512     & 71.86                      \\ %\cline{2-3} 
			& 256$\times$256     & 73.26                      \\ %\cline{2-3} 
			& 128$\times$128     & 74.96                      \\ %\cline{2-3} 
			& \cellcolor[HTML]{EFEFEF}\textbf{64$\times$64}       &   \cellcolor[HTML]{EFEFEF}\textbf{76.02} \\ %\cline{2-3} 
			& 32$\times$32       & \multicolumn{1}{c}{75.31} \\ \hline
	\end{tabular}}
	
\end{minipage}
\hspace{2mm}
\begin{minipage}[h]{0.47\textwidth}
	\centering
	\makeatletter\def\@captype{table}
	\scriptsize
	\setlength\tabcolsep{12pt}
	\renewcommand\arraystretch{1.4}
	\caption{Ablation studies of sample assignment strategy.}\label{table8}
	\vspace{-2mm}
	\resizebox{0.98\textwidth}{!}
	{
		\begin{tabular}{c|c|c}
			\hline
			Method                                & Lable Assignment & mAP   \\ \hline
			\multirow{2}{*}{Oriented R-CNN + OAN} & \cellcolor[HTML]{EFEFEF}\textbf{Center}           & \cellcolor[HTML]{EFEFEF}\textbf{76.02} \\ \cline{2-3} 
			& IoF              & 71.90 \\ \hline
	\end{tabular}}
	\vspace{0.9mm}
	\makeatletter\def\@captype{table}
	\scriptsize
	\setlength\tabcolsep{2pt}
	\renewcommand\arraystretch{1.3}
	\caption{Ablation studies of removing patch ratios.}\label{table9}
	\vspace{1mm}
	\resizebox{0.98\textwidth}{!}
	{
		\begin{tabular}{c|ccc|cc}
			\hline
			Method                              & Ratios  & Precision & Recall  & mAP                        & FPS                       \\ \hline
			Oriented R-CNN\cite{orcnn_iccv21}                      & -                          & -         & -       & 73.86                      & 15.6                      \\ \hline
			\multirow{9}{*}{\textbf{Oriented R-CNN + OAN}} & 0                          & 100\%     & 100\%   & \multicolumn{1}{c}{74.30} & \multicolumn{1}{c}{15.6}     \\ %\cline{2-6} 
			& 10\%                       & 98.68\%   & 99.99\% & 74.30                      & 16.7                      \\ %\cline{2-6} 
			& 20\%                       & 96.98\%   & 99.91\% & 74.31                      & 17.9                      \\ %\cline{2-6} 
			& 30\%                       & 94.04\%   & 99.74\% & 74.01                      & 19.6                      \\ %\cline{2-6} 
			& \cellcolor[HTML]{EFEFEF}\textbf{35\%}                       & \cellcolor[HTML]{EFEFEF}\textbf{89.71\%}   & \cellcolor[HTML]{EFEFEF}\textbf{99.36\%} & \cellcolor[HTML]{EFEFEF}\textbf{73.66} & \cellcolor[HTML]{EFEFEF}\textbf{20.4} \\ %\cline{2-6} 
			& 40\%                       & 85.18\%     & 98.83\% & 73.17                      & 21.3                      \\ %\cline{2-6} 
			& 45\%                       & 80.04\%     & 98.10\%   & \multicolumn{1}{c}{71.44} & \multicolumn{1}{c}{22.1} \\ %\cline{2-6} 
			& 50\%                       & 75.84\%   & 97.25\% & 68.96                      & \multicolumn{1}{c}{23.3} \\ %\cline{2-6} 
			& 60\%                       & 66.86\%   & 94.49\% & \multicolumn{1}{c}{62.10} & \multicolumn{1}{c}{25.8} \\ \hline
	\end{tabular}}
\end{minipage}
\vspace{6mm}

\noindent dimensions. Thus, we first reshape the 16$\times$16$\times$256 features into 8$\times$8$\times$1024 dimensions and then use the reshaped features for prediction. When the grid size is 32$\times$32, we first reduce the channel dimension of C5 with a 3$\times$3 stride-1 convolutional layer, and then directly perform prediction. 
Table \ref{Fig7} shows the results of Oriented R-CNN with OAN when using different sizes of grids. We have the observations from Table \ref{table7}: with the decrease of grid sizes from 1024$\times$1024 (equal to no grid division), i.e., beginning to use grid-wise prediction, the detection accuracy is increasingly improved, which further verifies the effectiveness of grid-wise classification. Especially, when the grid size comes to 64$\times$64, that is, dividing the input patch into 16$\times$16 grids, OAN obtains the highest accuracy. 
Unlike traditional methods (e.g., $R^2$-CNN) using a naive binary classifier to judge whether each patch contains objects or not, our OAN divides each patch into a set of grids and adopts the maximum value of the scores of all grids as the objectness metric for each patch. Thus, OAN can avoid the features of small-sized objects being submerged in the background to a degree and achieves more accurate prediction of objectness.

\textbf{Which sample assignment strategy is better?} To better investigate the effects of label assignment, we compare the results of two sample assignment strategies: center assignment and Intersection-over-foreground (IoF) assignment. For the center assignment, we have already discussed it in Section \ref{section3-1}. The IoF assignment is described as follows: (i) the grid that has an IoF overlap higher than 0.5 with any ground-truth box is set as positive; (ii) if the IoF overlaps of  a grid with all ground-truth boxes are lower than 0.1, the grid is defined as negative; (iii) other grids are ignored.

Table \ref{table8} reports the results of different sample assignment strategies. As can be seen, the center assignment is far superior to the IoF assignment. We conjecture this is mainly because of the differences of the quantity of positive samples. The center assignment uses a simple and loose positive-negative definition. It can 
assign each object to a grid regardless of its size and provides the supervision for the objectness prediction of each object, thus making the network easier to learn discriminative feature representations. On the contrary, the IoF assignment has a stricter positive-negative definition than that of center assignment. For the IoF assignment, whose number of positive samples is fewer than the center assignment, which prevents OAN from receiving more supervision and furthermore influences the learning of strong and robust feature~\cite{ene_cvpr21}. 

\textbf{Why could OAN work?} Readers may wonder why OAN eliminates many patches, but the detection accuracy is still improved. Are there no wrong predictions in OAN? To dispel the puzzle, we run a large

\noindent number of experiments on DOTA-v1.0 validation set, and analyze the detailed effect of OAN. Table \ref{table9} presents the results under different ratios of removed patches. It is worth noting that the definitions of precision and recall in Table \ref{table9} are different from the common conditions.
Here, we formulate the precision of OAN as: \emph{precision}= $\frac{\emph{\#correctly   filtered   patches}}{\emph{\#filtered  patches}}$. The recall of OAN is defined as: \emph{recall}= 1- $\frac{\emph{\#objects in all filtered patches}}{\emph{\#objects}}$.

As seen in Table \ref{table9}, we find that when we remove 30\% patches, the recall still achieves 99.74\% and the detection accuracy (mAP) is also better than the baseline, even if the error rate of OAN is 5.96\%. Firstly, we argue that this is because the incorrectly filtered patches contain very few objects. Secondly, these objects are difficult to be detected even if they pass to the detection network. Thirdly, the benefits of multi-task training and the reduction of false alarms (discussed in Section \ref{section5-4}) can balance the impact of the OAN's few wrong predictions to some extent. Thus we can improve the inference speed while obtaining slight mAP gain, under the error rate of 5.96\% (94.04\% precision).

In Figure \ref{Fig8}, we show all 172 patches (14\% of the total 1224 patches) activated with Eq. (\ref{equ6}) on a Gaofen-2 image and two zoomed-in patches with the objectness activation maps and corresponding detection results. As shown, the activation values of the areas with objects are larger. Meanwhile, the activation maps can better reflect the locations  of objects. This suggests that we can further use this information to dynamically estimate the number of proposals or infer approximate object locations.

In addition, one may wonder that the detectors has stronger supervision than OAN during training but why OAN can help detectors reduce false alarms? We argue that OAN does just use the supervision information of object \emph{vs.} non-object for grid-wise prediction, but it is confident for judging whether a patch contains objects. In this way, OAN can filter the patches without objects by adaptive threshold \emph{T} to let the detectors avoid outputting results on these patches. As a contrast, although the detectors leverage both the box and class information for supervision, according to common practice, the detectors will remain all the boxes whose scores are higher than 0.05 for achieving high recall. Since 0.05 is a very low threshold, the patches without any object still output some boxes, i.e., false positives. So, our OAN can reduce the false positives to some extent (see Figure \ref{Fig7}).

\begin{figure}[t]
\begin{minipage}[]{0.52\textwidth}
	\centering
	\makeatletter\def\@captype{table}
	\scriptsize
	\setlength\tabcolsep{23pt}
	\renewcommand\arraystretch{1.1}
	%\vspace{-5mm}
	\caption{Comparisons of OAN and its competitors.}\label{table10}
	\vspace{-1.1mm}
	\resizebox{0.95\textwidth}{!}
	{   \hspace{1mm}
		\begin{tabular}{l|cc}
			\hline
			\rowcolor[HTML]{FFFFFF} 
			& mAP   & FPS  \\ \hline
			\rowcolor[HTML]{FFFFFF} 
			ClusDet\cite{clusdet_iccv19}              & 47.60 & 10.2 \\
			\rowcolor[HTML]{FFFFFF} 
			DMNet\cite{dmnet_cvprw20}                & 60.80 & 12.1 \\
			\rowcolor[HTML]{FFFFFF} 
			$R^2$-CNN \cite{r2cnn_tgrs19}              & 66.20 & 17.8 \\ \hline
			\rowcolor[HTML]{EFEFEF} 
			\textbf{OAN} &\textbf{70.30}  & \textbf{18.5} \\ \hline
	\end{tabular}}
    \vspace{2mm}
    \centering
	\scriptsize
	\setlength\tabcolsep{3pt}
	\renewcommand\arraystretch{1.4}
	\caption{Ablation studies of different backbones.}\label{table11}
	\vspace{1.5mm}
	\resizebox{0.93\textwidth}{!}
	{
		\begin{tabular}{l|c|cc}
			\hline
			\rowcolor[HTML]{FFFFFF} 
			& Backbone  & mAP   & FPS  \\ \hline
			\rowcolor[HTML]{FFFFFF} 
			\cellcolor[HTML]{FFFFFF}                                       & ResNeSt50 & 76.22 & 13.2 \\
			\rowcolor[HTML]{FFFFFF} 
			\multirow{-2}{*}{\cellcolor[HTML]{FFFFFF}Oriented R-CNN\cite{orcnn_iccv21}}       & Swin-T    & 75.45 & 11.1 \\ \hline
			\rowcolor[HTML]{EFEFEF} 
			\cellcolor[HTML]{EFEFEF}                                       & ResNeSt50 & 76.52(+0.30) & 17.1(\textcolor{blue}{\textbf{+3.9}}) \\
			\rowcolor[HTML]{EFEFEF} 
			\multirow{-2}{*}{\cellcolor[HTML]{EFEFEF}\textbf{Oriented R-CNN + OAN}} & Swin-T    & 75.89(+0.44) & 14.5(\textcolor{blue}{\textbf{+3.4}}) \\ \hline
	\end{tabular}}
\end{minipage}
\hspace{4mm}
\begin{minipage}[]{0.41\textwidth}
	\centering
	%\makeatletter\def\@captype{table}
	
	\begin{center}
		\vspace{2mm}
		\hspace{0.5mm}\includegraphics[width=0.94\linewidth]{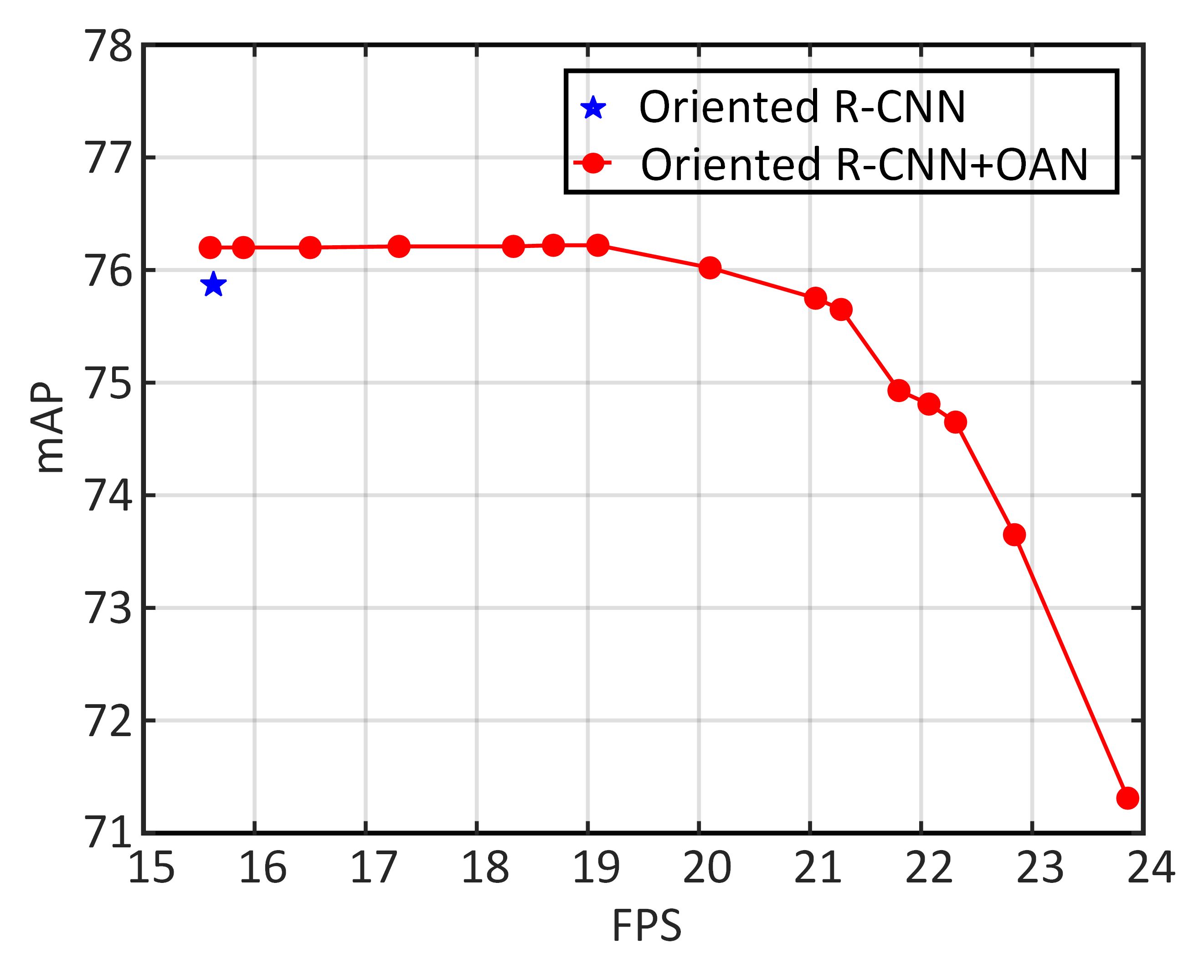}\\
	\end{center}
    \vspace{-8mm}
	\caption{The speed-accuracy trade-off curve.}\label{Fig9}
\end{minipage}
\vspace{2mm}
\end{figure}

\subsubsection{Comparisons with Competitors} 
To investigate the behaviors of OAN as a general and simple solution to speeding up the detection in large images, we compare OAN with three representative approaches, including $R^2$-CNN\cite{r2cnn_tgrs19}, ClusDet\cite{clusdet_iccv19} and DMNet\cite{dmnet_cvprw20}, on DOTA-v1.0 validation set.  We use Faster R-CNN as detector and the backbone is ReNet50-FPN. The settings of ClusDet and DMNet follow  their official codes.  For  $R^2$-CNN, we reproduce it because there is not publicly available source code. In Table \ref{table10}, we show the comparison results. As can be seen: (i) Faster R-CNN with OAN runs at 18.5 FPS, which is 81.3\% and 52.8\% faster than ClusDet and DMNet, respectively. (ii) Faster R-CNN with OAN obtains the best accuracy of 70.30\% mAP among all four methods, significantly outperforming ClusDet and DMNet by 22.70\% mAP and 9.50\% mAP, respectively. (iii) Compared with the most related work $R^2$-CNN, our OAN is leading in terms of accuracy (70.30\% mAP \emph{vs.} 66.20\% mAP) while keeping the advantage of inference speed.  

We argue that the vast superiority of OAN in terms of speed and accuracy against ClusDet and DMNet is caused by two reasons. The first reason is that OAN works in a light-weight full-convolutional fashion without time-consuming operations (e.g., region clustering and density mask generation) like ClusDet and DMNet. The other reason is that OAN is more simpler than ClusDet and DMNet, enabling network easier to learn and optimized end-to-end. As for $R^2$-CNN, the advantages of OAN with respect to  speed
\begin{figure*}[h]
	\begin{center}
		\includegraphics[width=0.98\linewidth]{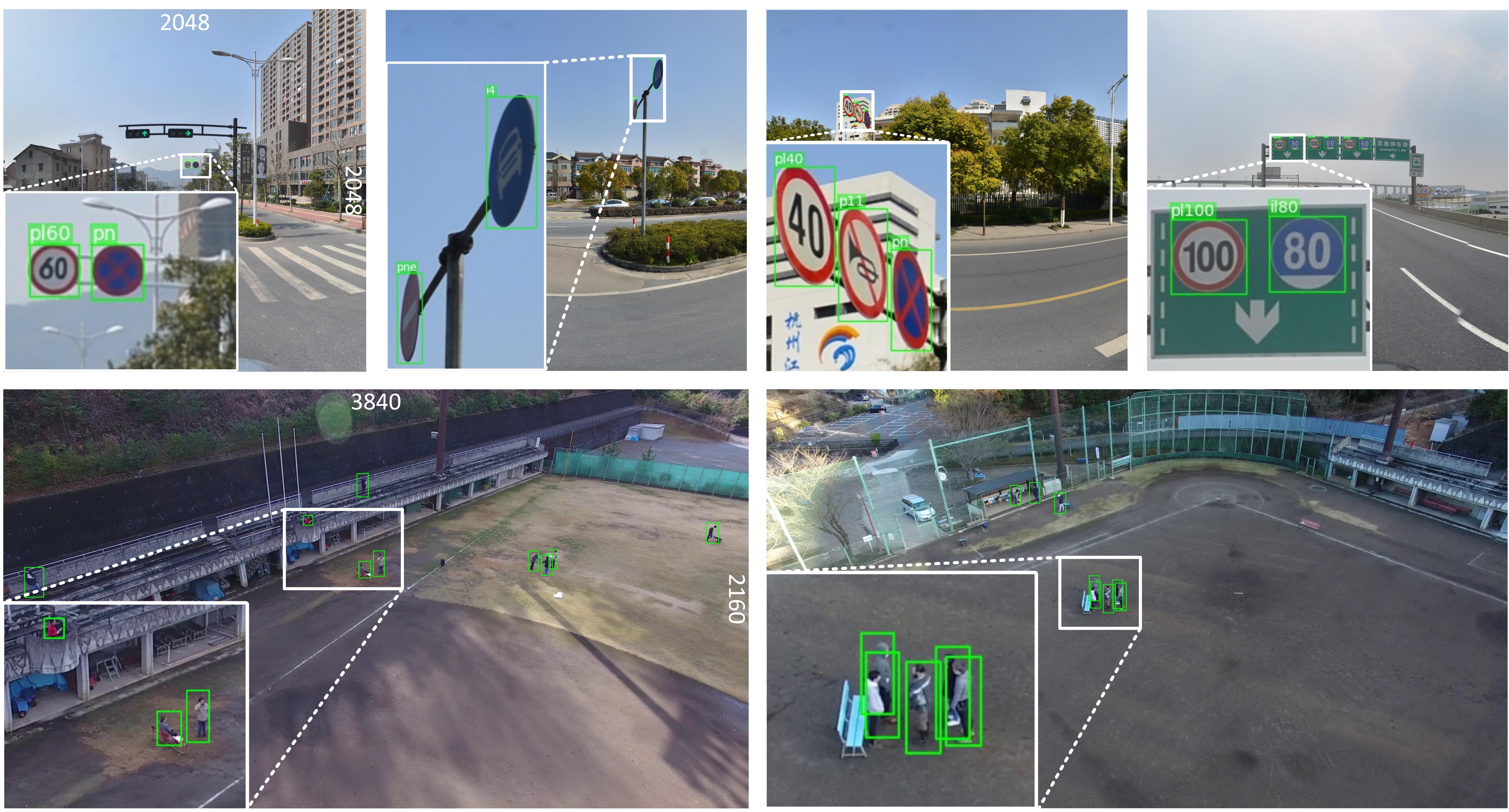}\\
	\end{center}
	\vspace*{-3mm}
	\caption{ Examples of detection results on the driving-scene dataset TT100K (top row) and 4K video dataset Okutama-Action (bottom row). Here, the symbols (top row) denote different traffic-sign classes, whose details can be found in \cite{tt100k_cvpr16}.}\label{Fig10}
	\vspace{2mm}
\end{figure*}

 \noindent and accuracy are mainly due to the following two aspects: (i) OAN implements objectness prediction by a fully-convolutional network and $R^2$-CNN performs objectness prediction via fully-connected layers. Therefore, OAN has fewer parameters than $R^2$-CNN (4.62M \emph{vs.} 4.75M), enabling faster speed. Here, the number of parameters of OAN and $R^2$-CNN are measured by counting their speed-up networks' parameters, using the calculation tool available from MMDetection~\cite{mmdetection_arxiv19}. (ii) OAN adopts grid-wise prediction that can better remain objects than $R^2$-CNN when the image patch only has fewer small-sized objects, thus achieving promising results. All these demonstrate OAN's simplicity and effectiveness again. 

\subsubsection{Trade-off between Speed and Accuracy}
OAN can greatly avoid numerous \emph{invalid patches} passing to the detectors, thus making detection more efficient. By changing the activation threshold $T$, we can control the number of patches passing through the whole detection network. With the increase of $T$, the inference becomes faster, and the detection accuracy improves to some extent but begins to decline when $T$ exceeds the limit. 
Specifically, in Figure \ref{Fig9}, we plot the speed-accuracy trade-off curve for Oriented R-CNN with OAN on DOTA-v1.0 test set. The backbone is R50-FPN. As shown, OAN significantly improves the speed from 15.6 FPS to more than 20 FPS, but its accuracy is still higher than the baseline.

\subsubsection{Robustness to Different Backbones}
OAN shares backbones with the detectors. How do different backbones influence the accuracy and speed of detection? To verify the robustness of OAN to different advanced backbones, we experiment with two state-of-the-art backbones, namely ResNeSt50\cite{resnest_arxiv2020} and Swin-T\cite{swin_iccv21}, which are based on convolutional structure and transformer, respectively. In Table \ref{table11}, we show the performance improvements obtained from OAN when using ResNeSt50 and Swin-T as backbones, respectively. The results clearly present that OAN can consistently improve the accuracy and speed with different baselines. For example, when using Swin-T, applying OAN to Oriented R-CNN accelerates the inference speed by 30.6\% and meanwhile obtains a gain of 0.44\% mAP. This suggests that our OAN is robust to different backbones. We believe the robustness of OAN  will benefit the use of OAN on various advanced backbones.

\begin{figure}
\hspace{0.1mm}
	\begin{minipage}[]{0.49\textwidth}
		\makeatletter\def\@captype{table}
		\scriptsize
		\setlength\tabcolsep{2.2pt}
		\renewcommand\arraystretch{1.5}
		\caption{Driving-scene object detection results on the TT100K dataset. }\label{table12}
		\resizebox{1\textwidth}{!}
		{
			
			\begin{tabular}{l|ccc|c}
				
				\hline
				& AP    & $\text{AP}_{50}$ & $\text{AP}_{75}$ & FPS  \\ \hline
				Faster R-CNN\cite{faster_tpami17}       & 86.42 & 50.62    & 50.81    & 15.6 \\
				RetinaNet\cite{retinanet_iccv17}          & 78.31 & 45.86    & 46.91    & 16.9 \\ 
				\rowcolor[HTML]{EFEFEF} \hline
				\textbf{Faster R-CNN + OAN} & 86.69(+0.27) & 51.71    & 52.43    & 27.3(\textcolor{blue}{\textbf{+11.7}}) \\
				\rowcolor[HTML]{EFEFEF} 
				\textbf{RetinaNet + OAN}    & 78.91(+0.60) & 48.20    & 48.52    & 28.8(\textcolor{blue}{\textbf{+11.9}}) \\ \hline
			\end{tabular}
		}
	\end{minipage}
	\hspace{1.1mm}
	\begin{minipage}[]{0.47\textwidth}
		
		\makeatletter\def\@captype{table}
		\scriptsize
		\setlength\tabcolsep{2pt}
		\renewcommand\arraystretch{1.54}
		\caption{4K Video object detection results on the Okutama-Action dataset.}\label{table13}
		%\vspace{-2mm}
		\resizebox{1\textwidth}{!}
		{
			\begin{tabular}{l|ccc|c}
				\hline
				& AP    & $\text{AP}_{50}$ & $\text{AP}_{75}$ & FPS  \\ \hline
				Faster R-CNN\cite{faster_tpami17}       & 86.42 & 50.62    & 50.81    & 15.6 \\
				RetinaNet\cite{retinanet_iccv17}          & 78.31 & 45.86    & 46.91    & 16.9 \\ 
				\rowcolor[HTML]{EFEFEF} \hline
				\textbf{Faster R-CNN + OAN} & 86.69(+0.27) & 51.71    & 52.43    & 27.3(\textcolor{blue}{\textbf{+11.7}}) \\
				\rowcolor[HTML]{EFEFEF} 
				\textbf{RetinaNet + OAN}    & 78.91(+0.60) & 48.20    & 48.52    & 28.8(\textcolor{blue}{\textbf{+11.9}}) \\ \hline
		\end{tabular}}
	\end{minipage}
	\vspace{2mm}
\end{figure}
 
\noindent

\section{Extended Experiments }
Beyond object detection in large aerial images, we extend our OAN to some specific object detection tasks, like driving-scene object detection and 4K video object detection. Two popular detectors, termed Faster R-CNN and RetinaNet, are used as baselines for both the tasks, and the backbone is ResNet50. As same as the setting in Section \ref{section5-1}, we divide each large image into patches with the size of 1024$\times$1024, and the overlap between adjacent patches is 200 pixels. AP, $\text{AP}_{50}$, $\text{AP}_{75}$, and FPS are used  for evaluation metrics. Here, the AP is the average AP across the IoU thresholds from 0.5 to 0.95 with 0.05 interval. Next, we will use OAN to speed up driving-scene object detection and 4K video object detection, respectively.

\subsection{OAN for Driving-scene Object Detection}
Driving-scene object detection plays an important role in understanding the overall scenes for autonomous driving planning and execution{\cite{kitti_cvpr12}}. In autonomous driving scenes, the images, captured by vehicle cameras, always have high resolution. So it is meaningful for autonomous driving to achieve efficient object detection on vehicle devices with limited capacity of computation and memory. To demonstrate the generality and practical value of our OAN, we test OAN on the large-scale traffic-sign detection dataset, termed TT100K{\cite{tt100k_cvpr16}}, which contains 45 traffic-sign classes, and 100000 driving-scene images (2048$\times$2048 pixels) with large variations in illumination and weather conditions in the wild. Following the previous works, 10000 images that have traffic-signs are divided into training set and testing set with 2:1 ratio, and the remaining 90000 images without objects are used for testing set. 

In Table \ref{table12}, we report the results of our methods and the baselines (Faster R-CNN and RetinaNet). As seen in Table \ref{table12}, OAN can significantly improve the detection speed of baselines. Firstly, with OAN, Faster R-CNN and RetinaNet achieve more than 112.1\% and 98.8\% speed-up, respectively. Secondly, OAN still leads to consistent improvements of accuracy  under stricter metrics. These results are consistent with previous experiments and further prove the generality of OAN. In addition, in the top row of Figure \ref{Fig10}, we show some detection results on the TT100K dataset by using Faster R-CNN with OAN.

\subsection{OAN for 4K Video Object Detection}
With the advance in smart devices (cell phones, TVs and cameras), more and more 4K even 8K videos have entered our daily life. How to achieve efficient processing for 4K or 8K videos has become more and more important. To further verify OAN's generalization ability and push it to enable real-world applications, we extend OAN to 4K video object detection \cite{vod_cvpr18} and use a challenging video dataset, termed Okutama-Action, for experiments.  Okutama-Action{\cite{okutama_cvprw17}} is designed for pedestrian detection and human action detection. It consists of 43 video sequences with  4K  resolution and covers 12 action classes. 33 video sequences belong to training and the remaining for testing. All videos are captured from UAVs flying at different altitudes and angles. Since the focus of our OAN is efficient object detection, we only test OAN on the pedestrian class without considering the task of human action detection. 

In Table \ref{table13}, we present the performance of our method and the baselines. The baselines, Faster R-CNN and RetinaNet, have 86.42\% AP and 78.31\% AP and run at 15.6 FPS and 16.9 FPS, respectively. Using OAN, Faster R-CNN and RetinaNet could  acquire more than 75.0\% and 70.4\%\ speed-up as well as obtaining better accuracy. This indicates that our OAN performs excellently for improving the detection speed in the task of 4K video object detection. In the bottom row of Figure \ref{Fig10}, we visualize some detection results. We hope such a big improvement of speed will benefit more tasks of very high-resolution video processing, such as object tracking and video segmentation.

\section{Conclusion}
In this work, we proposed a simple but effective objectness activation network (OAN) used for judging whether there exist objects in input patches. OAN is a light fully-convolutional network, which adopts grid-wise prediction and uses the maximum value of the scores of all grids as the objectness metric for each patch. It can be applied to many object detectors, and jointly optimized with them end-to-end, enabling a simple, effective, and unified detection framework for large images. Using OAN, we can improve the inference speed of detectors by more than 30.0\%, while achieving consistent improvements of accuracy. Besides, for extremely large images, we got 70.5\% speed improvement without sacrificing accuracy.  Moreover, we extended our OAN to driving-scene object detection and 4K video object detection, which improves the inference speed by 112.1\% and 75.0\% without sacrificing accuracy, respectively. We hope the strong performance of OAN will encourage the research on efficient object detection in large-size images and facilitate real applications.

In addition, the limitation of our OAN is that though we achieve more than 30.0\% speed-up for mainstream detectors, abundant information in OAN's grid-wise activation maps is not fully leveraged. We will utilize this information to furthermore improve detection performance in the future work.

%%%%%%%%%%%%%%%%%%%%%%%%%%%%%%%%%%%%%%%%%%%%%%%%%%%%%%%
%%% Acknowledgements. ÖÂÐ»
%%%%%%%%%%%%%%%%%%%%%%%%%%%%%%%%%%%%%%%%%%%%%%%%%%%%%%%
\Acknowledgements{This work was supported in part by the National Natural Science Foundation of China under Grants 62136007 and U20B2065, in part by the Natural Science Basic Research Program of Shaanxi under Grants 2021JC-16 and 2023-JC-ZD-36, in part by the Fundamental Research Funds for the Central Universities, and in part by the Doctorate Foundation of Northwestern Polytechnical University under Grant CX2021082.}

%%%%%%%%%%%%%%%%%%%%%%%%%%%%%%%%%%%%%%%%%%%%%%%%%%%%%%%

%%%%%%%%%%%%%%%%%%%%%%%%%%%%%%%%%%%%%%%%%%%%%%%%%%%%%%%
%%% Appendix sections. ¸½Â¼ÕÂ½Ú, ·Ç±ØÑ¡
%%%%%%%%%%%%%%%%%%%%%%%%%%%%%%%%%%%%%%%%%%%%%%%%%%%%%%%
%\begin{appendix}
%\section{Name}

%\end{appendix}

\end{document}